\documentclass[screen,acmtog]{acmart}

\copyrightyear{2022} 
\acmYear{2022} 
\setcopyright{acmcopyright}\acmConference[KDD '22]{Proceedings of the 28th ACM SIGKDD Conference on Knowledge Discovery and Data Mining}{August 14--18, 2022}{Washington, DC, USA}
\acmBooktitle{Proceedings of the 28th ACM SIGKDD Conference on Knowledge Discovery and Data Mining (KDD '22), August 14--18, 2022, Washington, DC, USA}
\acmPrice{15.00}
\acmDOI{10.1145/3534678.3539059}
\acmISBN{978-1-4503-9385-0/22/08}

\newcommand{\eat}[1]{}
\newcommand{\hush}[1]{}

\usepackage{amsmath}
\usepackage{amsmath,amsfonts}
\usepackage{algorithmic}
\usepackage{algorithm}
\usepackage{bm}
\usepackage{graphicx}
\usepackage{textcomp}
\usepackage{xcolor}
\usepackage{cuted, nccmath}
\usepackage{lipsum}
\usepackage{mathtools}
\usepackage{graphicx}
\usepackage{xspace}
\usepackage{microtype}
\usepackage{subfigure}
\usepackage{booktabs} % for professional tables

\usepackage{url}

\usepackage{breakurl}

\usepackage{soul} % for hl

\usepackage{hyperref}

\newcommand{\Looper}{Looper\xspace}
\newcommand{\Meta}{Meta\xspace}
\newcommand{\Facebook}{Facebook\xspace}
\newcommand{\Instagram}{Instagram\xspace}
\newcommand{\anoncite}[1]{~\cite{#1}}

\usepackage{tikz}
\newcommand*\circled[1]{\tikz[baseline=(char.base)]{
            \node[shape=circle,draw,inner sep=2pt] (char) {#1};}}

\usepackage{xspace}

\begin{document}

\title{Looper: an end-to-end ML platform for product decisions}

\author{Igor Markov}
\email{imarkov@fb.com}
\author{Hanson Wang} % hanson.wng@gmail.com
\email{hanson.wng@gmail.com}
\author{Nitya Kasturi}
\author{Shaun Singh}{} % shaundsingh@gmail.com
\author{Mia Garrard}
\author{Yin Huang}
\author{Sze Wai Yuen}
\author{Sarah Tran}
\author{Zehui Wang} % zehuiw.wang@gmail.com
\author{Igor Glotov}
\author{Tanvi Gupta}
\author{Peng Chen}
\author{Boshuang Huang}
\author{Xiaowen Xie} % {google} xwx@google.com
\author{Michael Belkin}
\author{Sal Uryasev}
\author{Sam Howie}
\author{Eytan Bakshy}
\author{Norm Zhou}
\affiliation{%
  \institution{Meta}
  % \institution{Meta AI Platforms and Meta Core Data Science}
  \streetaddress{One Hacker Way}
  \city{Menlo Park}
  \state{CA}
  \country{USA}
  \postcode{94025}
}
\email{nzhou@fb.com}
%\email{imarkov,nityakasturi,hansonw,shaunsingh,szewaiyuen6,sarahtran,mgarrard,wzehui,maggiehuang,hbs,igorglotov,tanvigupta,pengchen18,showie,ebakshy,nzhou @fb.com}

\renewcommand{\shortauthors}{Markov and Wang, et al.}

%\mlsysaffiliation{fb}{Meta, Menlo Park, CA}

%\mlsyscorrespondingauthor{Igor L. Markov}{imarkov@fb.com}
%\mlsyscorrespondingauthor{Norm Zhou}{nzhou@fb.com}

%\mlsysaffiliation{fb}{\institution{Facebook Inc., Menlo Park, CA}}

%\email{imarkov,nityakasturi,hansonw,shaunsingh,szewaiyuen6,sarahtran,mgarrard,wzehui,maggiehuang,hbs}
%\email{igorglotov,tanvigupta,pengchen18,showie,ebakshy,nzhou 
%@fb.com};

%\vskip 0.2in

\begin{abstract}
  Modern software systems and products increasingly rely on machine learning models to make data-driven decisions based on interactions with users, infrastructure and other systems. For broader adoption, this practice must ($i$) accommodate product engineers without ML backgrounds, ($ii$) support finegrain product-metric evaluation and ($iii$) optimize for product goals. To address shortcomings of prior platforms, we introduce general principles for and the architecture of an ML platform, \textit{\Looper}, with simple APIs for decision-making and feedback collection. 
  \Looper covers the end-to-end ML lifecycle from collecting training data and model training to deployment and inference, and extends support to personalization, causal evaluation with heterogenous treatment effects, and Bayesian tuning for product goals. During the 2021 production deployment, \Looper simultaneously hosted 440-1,000 ML models that made 4-6 million real-time decisions per second. We sum up experiences of platform adopters
  and describe their learning curve.%
\end{abstract}




\maketitle

%\printAffiliationsAndNotice{}   % otherwise use the standard text.

\section{Introduction}
\label{sec:intro}

\hush{
The economics of software development justifies high capital investment with large user bases, e.g., for e-commerce, search engines, social networks, gaming and entertainment, online advertising, financial and medical services.  Interactive front-ends power user experiences with growing complexity, and rely on remote back-ends for account maintenance, data access and persistence, as well as data interpretation~\cite{rudman2016defining}.}

With growing adoption of machine learning (ML), personalization is proving essential to competitive user experience~\cite{marc_darcy_2021}. To support users with different preferences, one needs good default tactics, user feedback, prioritizing delivered content and available actions~\cite{molino21declarative}.
%When managing limited resources, similar logic applies to network bandwidth, response latency and screen space~\cite{letham2019constrained,mao2020real}.
When managing limited resources, e.g., for video serving, similar logic applies to network bandwidth, response latency, and video quality~\cite{mao2020real, feng2020high}.
This paper explores the use of ML for personalized decision-making in software products using what we call \textit{smart strategies} (Section \ref{sec:smart_strat_ml}). Making smart strategies
available to product engineers is challenging\eat{~\cite{agarwal2016making,molino21declarative}}:
 \circled{1} Long-term product objectives rarely match closed-form ML loss functions.
 \circled{2} Product-generated data drifts away from training data.
\circled{3} Capturing correlations in training data (via ML) does not imply causal improvement of product metrics~\cite{bakshy2014designing,xu2015infrastructure}.
\circled{4} Clean data to evaluate the performance of ML systems is often unavailable, necessitating A/B testing.
\circled{5} Traditional A/B tests neglect personalized treatments.
\eat{~\cite{letham2019policy}} 
\circled{6} Platforms to train, host and monitor hundreds of ML models are needed and promise economies of scale.
\circled{7} Real-time feature extraction and inference are needed, despite more efficient async batch processing.
\circled{8} Product engineers, many new to ML, need a simple, standard, future-proof way to embed smart strategies into products.

% Below we outline these challenges and how they can be addressed.

\noindent
{\bf Data-centric ML development} is a recent concept of refocusing ML development from models to data~\cite{miranda2021datacentric}. It supports software personalization with off-the-shelf models, where collecting the right data and selecting the appropriate class of models become primary differentiators~\cite{molino21declarative}.\eat{Aside from traditional data management concerns, ML systems for personalization struggle to handle the noise inherent in user feedback signals and product impact metrics~\cite{letham2019constrained}.}
\hush{the scalability of feature management;}\eat{It is challenging to select features relevant to the task from a sea of available features with different computational cost profiles.}%
Compared to developing and training ML models, data adequacy is often overlooked~\cite{sambasivan2021everyone}, and product platforms must use automation to compensate. Per Andrew Ng, ``everyone jokes that ML is 80\% data preparation, but no one seems to care''~\cite{sagar_2021}.
\eat{Data and model quality aside, product decisions are driven by {\em product goals}, and impact on numerous users is measured via A/B tests ~\cite{bakshy2014designing,xu2015infrastructure,letham2019policy}.}
Yet, directly handling data sets and ML models in product code is cumbersome. Instead, \textit{software-centric} ML integration with data collection and decision-making APIs offers a front-end to MLOps automation (Sections \ref{sec:end2end} and \ref{sec:core}). \hush{On the product-metric side}
Additionally, ML development often neglects structure in product evaluation data (Section \ref{sec:special}).
\eat{
Scaling, productionizing and fully measuring the impact of smart strategies calls for \textit{software-centric} ML integration with APIs for data collection and decision-making, rather than application code directly dealing with models and data sets.
}

\begin{figure*}[t]
  \centering
  \vspace{-2.6mm} 
  \includegraphics[trim=0 4mm 0 0, clip,width=0.70\linewidth]{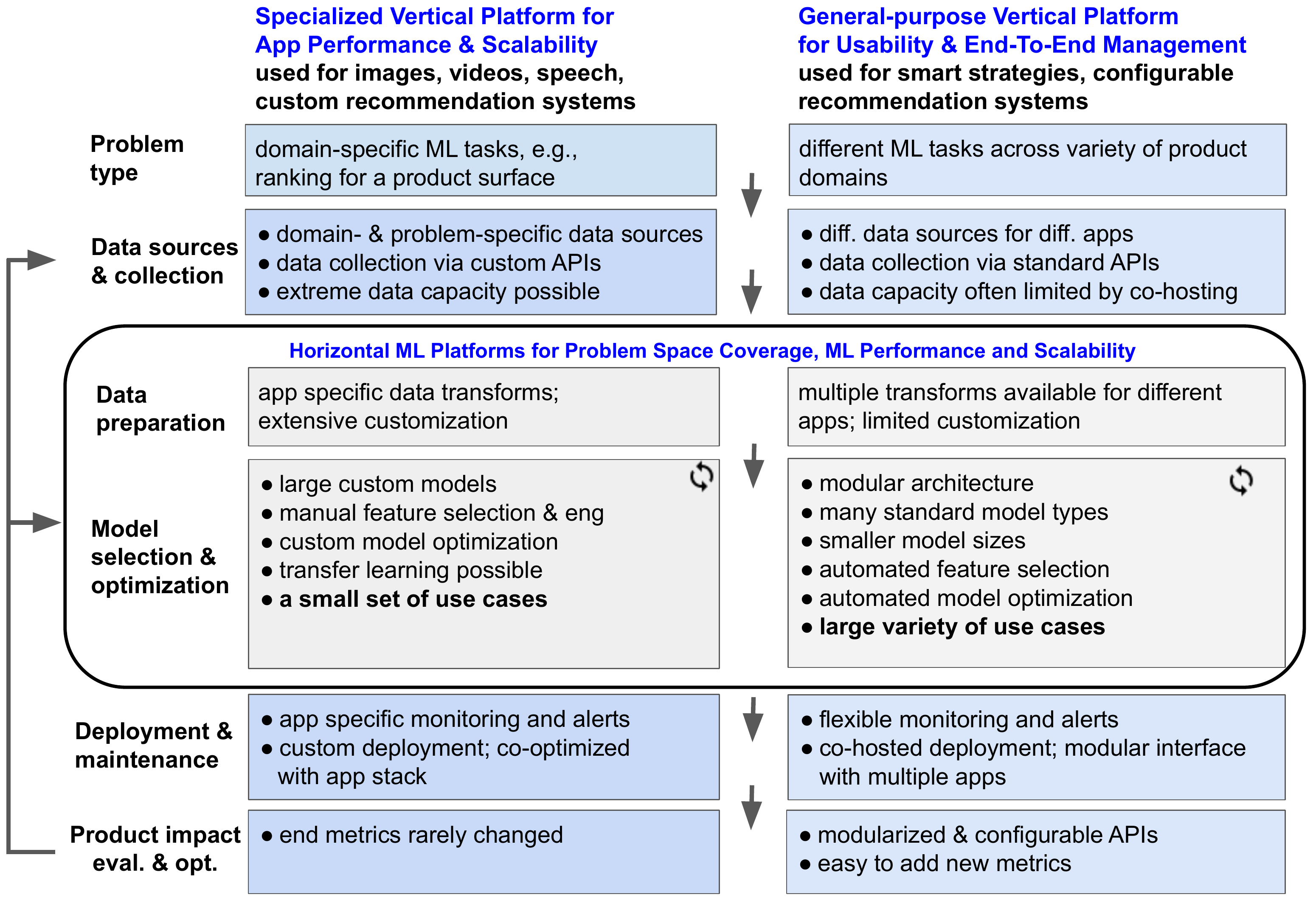}
  \vspace{-4mm}
  \parbox{16cm}{
  \caption{\label{fig:platforms}
  Categories of applied ML platforms: horizontal vs. vertical, specialized vs. general-purpose (back arrows show vertical optimizations based on product metrics, see Section \ref{sec:blueprints}). Specialized platforms are limited in their support for diverse applications.
  }
 }
  % https://docs.google.com/presentation/d/1O3kmMp_xDb7wnqamoFVq34CcYCfy8kQhUqbWvH0pivE/edit?usp=sharing
  %\vspace{-1mm}
\end{figure*}

\noindent
{\bf Vertical ML platforms} lower barriers to entry and support the entire lifecycle of ML models (Figure \ref{fig:platforms}) in a repeatable way. Horizontal ML platforms provide storage, support data pipelines and offer basic services, whereas vertical platforms foster the reuse of not only ML components, but also workflows.
% orchestrate available modules and resources for specific application needs, including development, productization, evaluation, operation and maintenance activities 
At firms like Google, Meta, LinkedIn, Netflix,
{\em specialized end-to-end vertical platforms} drive flagship product functionalities, such as recommendations. They have also been applied to software development, code quality checks, and even to optimize algorithms such as sorting and searching~\cite{carbune2018smartchoices}.
Platforms are built on ML frameworks like TensorFlow \cite{tensorflow2016} and PyTorch \cite{pytorch2020} that focus on modeling for generic ML tasks, support hardware accelerators, and act as toolboxes for application development \cite{gauci2018horizon,molino19ludwig}.
Supporting smart strategies requires {\em general-purpose vertical platforms} to offer end-to-end ML lifecycle management. General-purpose vertical ML platforms can be internal to a company --- Apple’s Overton \cite{re2019overton} and Uber's Michelangelo \cite{Hermann2017Michelangelo}, --- or
broadly available to cloud customers ---
Google's Vertex, Microsoft’s Azure Personalizer \cite{agarwal2016making} and
Amazon Personalize. A common theme
is to help engineers
``build and deploy deep-learning applications without writing code” via high-level, declarative abstractions \cite{molino21declarative}.
Improving user experience and system performance with ML remains challenging~\cite{paleyes2020challenges}
as correlations in data found by ML models might not lead to causal improvements.
Little is known about optimizing for product goals~\cite{molino21declarative, wu2021sustainable}. 

\hush{
Microsoft’s Azure Personalizer can be used “as a standalone personalization solution or to complement an existing solution—with no machine learning expertise required” \cite{agarwal2016making}. Amazon Personalize is linked with AWS and “enables developers to build applications with ... ML technology used by Amazon.com for real-time personalized recommendations – no ML expertise required.” 
\hush{This service is integrated with AWS and offers intelligent resource management: “You will receive results via an Application Programming Interface (API), and only pay for what you use… All data is encrypted to be private and secure, and is only used to create recommendations for your users”.}
Apple’s Overton \cite{re2019overton} helps engineers \hush{``in building, monitoring, and improving production machine learning systems, so that engineers can ''}
``build and deploy deep-learning applications without writing code” by supporting high-level, declarative abstractions based on schemas. \hush{Overton offers offline model tuning\hush{with fine-grained quality evaluation}, error diagnosis, and weak supervision to handle contradictory/incomplete labels.}
The Ludwig platform supports
``deep learning model building based on... data types and declarative configuration files"
\cite{molino19ludwig}. Uber's Michelangelo ``enables Uber’s product teams to seamlessly build, deploy, and operate machine learning solutions'' \cite{Hermann2017Michelangelo}. 
% Like many other platforms, Michelangelo emphasizes offline processing.
}

% Evaluating and ensuring product impact requires validation with A/B testing on live data ~\cite{bakshy2014designing, xu2015infrastructure,bakshy2018ae}, so as to guide system development by business or engineering impact, not abstract ML performance. 

%\noindent
We develop support for {\em data-driven real-time smart strategies} via a {\em general-purpose vertical end-to-end} ML platform called \textit{\Looper}, internal to \Meta, for
rapid, low-effort deployment of moderate-sized models. \Looper is a \textit{declarative ML system}  \cite{Hermann2017Michelangelo,molino19ludwig,re2019overton,molino21declarative}
with coding-free full-lifecycle management of smart strategies via a GUI.

\noindent
{\bf Our technical contributions} include
\circled{1}
a full-stack real-time ML platform (Section \ref{sec:Looper}) with causal product-impact evaluation and optimization and
handling of heterogeneous treatment effects (Sections \ref{sec:special}, \ref{sec:pex_impact})
via an experiment optimization system\hush{~\cite{bakshy2014designing,bakshy2018ae}} and meta-learners,
\circled{2}
 a generic framework for {\em targeting long-term outcomes} by parameterized policies using plug-in supervised learning models and Bayesian optimization (Sections \ref{sec:blueprints}, \ref{sec:special}, \ref{sec:pex_impact}),
\circled{3} the {\em strategy blueprint} abstraction\hush{And finally, to better manage the full lifecycle of smart strategies, we develop the concept of a {\em strategy blueprint} in Section \ref{sec:blueprints}.} to optimize not only models, but the entire ML stack (Figures \ref{fig:platforms}, \ref{fig:blueprint} and Section \ref{sec:blueprints}), 
\circled{4} 
capturing decision inputs and observations online via the succinct \Looper API for product code;\eat{This generalizes reinforcement learning APIs in Microsoft's Decision Service \cite{agarwal2016making} and Google's SmartChoices  \cite{carbune2018smartchoices}}
not only predicting what's logged by the API, but also optimizing black-box product objectives (Section \ref{sec:core}),
\eat{while
% Simplified data management 
preventing mismatches between training and deployment seen in offline-first platforms
}
%outcomes that can be directly predicted by %the underlying predictive models.
\hush{(Section \ref{sec:core}).} 
% The use of a similar API for software decision-making was introduced in SmartChoices \cite{carbune2018smartchoices}. However, the authors target self-contained algorithms, rather than entire production systems. They do not address MLOps complexities, personalization, or A/B testing. 
% In addition to capturing data and observations,  \Looper APIs have direct access to product metrics
%
\circled{5} broad deployment and
substantial impact on product metrics
(Section \ref{sec:impact}),
\circled{6} analysis of resource-usage bottlenecks
(Appendix \ref{sec:resources}),
\circled{7} qualitative analysis of our platform via a survey of adopters
(Appendix \ref{sec:survey}).

Specialized vertical ML platforms limit application diversity, while \Looper hosts hundreds of production use cases thanks to its {\em general-purpose architecture}.
Many vertical platforms~\cite{agarwal2016making,carbune2018smartchoices,re2019overton,molino19ludwig}, 
don’t solve as wide a selection of ML tasks as \Looper does (classification, estimation, value and sequence prediction, ranking, planning) using supervised and reinforcement learning.\hush{Combined with model management infrastructure,} 
Unlike platforms with asynchronous batch-mode feature extraction and inference \cite{Hermann2017Michelangelo,Gupta2020FBDNN}, \Looper runs in real time and optimizes resource usage accordingly (Appendix \ref{sec:resources}).
To balance model quality, size and inference time, \Looper AutoML selects models and hyperparams, and performs vertical optimizations via {\em strategy blueprints} (Section \ref{sec:blueprints}).
\hush{\Looper covers the scope from data sources to product impact, evaluated and optimized via causal experiments.}
% We report \Looper's impact on deployed products at \Meta and give common reasons supporting and blocking adoption.

\noindent
{\bf In the remainder of the paper},  Section~\ref{sec:smart_strat_ml} explores ML-driven smart strategies and relevant platform needs. Section~\ref{sec:Looper} covers design principles for the \Looper platform, introduces the architecture, the API, the blueprints, and specializations.
Section~\ref{sec:impact} summarizes product impact at \Meta, comparisons to baselines and adoption statistics. 
%Secton \ref{sec:conclusions} reviews how \Looper helps improve SW systems and products. 
Appendices provide data to foster reproduciibility.
\eat{Appendix~\ref{sec:survey} covers the adoption of smart strategies\eat{and the barriers to it}.}
\eat{via product-driven ML-based smart strategies.}
%and discuss the organizational support required.

\vspace{-2mm}
\section{ML for smart strategies}
\label{sec:smart_strat_ml}

Compared to benchmark-driven research, ML that interacts with the world runs into additional challenges. In this paper, we target smart strategies at key decision points in software products, e.g.,
\begin{itemize}
\vspace{-1mm}
\item 
application settings and preferences: selecting between defaults and user-specified preferences
%\item account and feature access, security response tactics
\item adaptive interfaces --- certain options are shown only to users who are likely to pursue them
\item controlling the frequency of ads, user notifications, etc
\item prefetching or precomputation to reduce \hush{data fetching} latency
\item content ranking and prioritizing available actions
%\item tradeoffs between competing product goals
%, e.g., resource costs vs user engagement.
\vspace{-1mm}
\end{itemize}

User preferences and context complicate decision-making. Simplifying a UI menu can boost product success, but menu preferences vary among users. Prefetching content to a mobile device can enhance user experience, but may require predicting user behavior.

\hush{Machine learning is naturally suited for applications where massive datasets defy simple patterns or engineers can’t develop robust rules or scalable algorithms, often because of diverse or changing data. To unlock likely gains, we ($i$) focus on the relevance and practical impact of ML models, ($ii$) use ML as a force multiplier for system development.}

While human-crafted heuristic strategies often suffice as an initial solution, ML-based smart strategies tend to outperform heuristics upon sufficient engineering investment \cite{kraska2017learned, carbune2018smartchoices}. The \Looper platform aims to lower this crossover point to broaden the adoption of smart strategies and deliver product impact over diverse applications.
In this section, we discuss modeling approaches to enable smart strategies and cover the priorities in building such a platform.

\subsection{Modeling approaches for smart strategies} 
\label{sec:blueprints_model}

Smart strategies are backed by supervised learning, contextual bandits (CB), and/or MDP-style reinforcement learning (RL). Given a model type, product decision problems need (1) ML optimization objectives to approximate the product goal(s) and (2) a decision policy to convert objective predictions into a single decision.

% have some key differences in data abstractions and joining procedures: (1) supervised-learning labels are associated directly with single decision points, whereas RL may aggregate long-term rewards over multiple decisions (2) long-term rewards in sequential RL are usually joined via offline data transform pipelines while short-term observations may be observed online.

\noindent
{\bf Approximating product goals with predictable outcomes} (alternatively referred to as \textit{proxy} or \textit{surrogate} objectives) is a major difference between industry practice and research driven by existing ML models with abstract optimization objectives~\cite{stein2019}. Good proxy objectives should be readily measurable and reasonably predictable. In recommendation systems, the “surrogate learning problem has an outsized importance on performance in A/B testing but is difficult to measure with offline experiments” \cite{covington2016deep}. 
We note a delicate tradeoff between easy-to-measure objectives directly linked to the decision vs. more complex objectives, e.g., ad clicks vs. conversions. 
Furthermore, product goals often implicitly have different weighting functions than the ML objective (e.g., the feedback provided by most prolific product users\hush{must be scaled down since it} does not always represent other users~\cite{beutel2017beyond}). 
Objectives can be modeled directly using \textit{supervised learning}; alternatively, using CBs can model uncertainty in predictions across one or more objectives, which may then be used for exploring the set of optimal actions, e.g., in Thompson sampling \cite{agarwal2009explore,li2010contextual,agarwal2016making,daulton2019thompson}. The use of RL enables the optimization of long-term, cumulative objectives, helping use cases with sequential dependencies~\cite{li2010contextual,gauci2018horizon,apostolopoulos2021personalization}. 
To evaluate any one of these types of models and decision rules, true effects of the ML-based smart strategies can be estimated via A/B tests.
% To match ML objectives with product metrics, product teams customize training data, penalty terms, and postprocessing.

\noindent
{\bf Decision policies} postprocess the raw model outputs into a final product  decision or action. For single-objective tasks in supervised learning this may be as simple as making a binary decision if the objective prediction exceeds a threshold, e.g. turning the probability of a click into a binary prefetch decision 
(Section \ref{sec:prefetch}). For tasks with multiple objectives and more complex action spaces, the template for a decision policy is to assign a scalar value or score to all possible \textit{actions} in the decision space, which can then be ranked through sorting. In recommendation systems, a standard approach is to use a combination function (usually a weighted product of objective predictions) to generate a score for each candidate \cite{zhao2019recommending}. When using reinforcement learning, reward shaping \cite{laud2004reward}  weighs task scores in the reward function to optimize for the true long-term objective. Optimizing this weighting for multi-objective tasks is explored in Section \ref{sec:blueprints}. More sophisticated policies also use randomization to explore the action space, e.g. Thompson sampling in contextual bandits \cite{daulton2019thompson}, or $\varepsilon$-greedy approaches for exploration in ranking \cite{agarwal2009explore}.

\subsection{Extending end-to-end ML for smart strategies}
\label{sec:end2end}
Traditional end-to-end ML systems go as far as to cover model publishing and serving~\cite{Hermann2017Michelangelo,molino19ludwig,re2019overton,molino21declarative}, but to our knowledge rarely track {\em how} the model is used in the software stack. Assessing and optimizing the impact of smart strategies, especially with respect to product goals, requires experimentation on all aspects of the modeling framework -- from metric and model selection to policy optimization. To streamline this experimentation and reap its benefits, smart-strategies platforms must extend the common definition of end-to-end into the software layer.
% End-to-end ML systems imply a different set of priorities compared to standalone ML models. The general lifecycle of an end-to-end ML solution involves \cite{paleyes2020challenges}: 
% \begin{itemize}
%     \item obtaining sufficient high-quality data to learn from, 
%     \item providing reliable evaluation of predictive models, 
%     \item ensuring robust product performance over time, and
%     \item facilitating sustainable resource usage.
% \end{itemize}

\noindent
{\bf Software-centric ML integration}  \cite{agarwal2016making,carbune2018smartchoices}
-- where data collection and decision-making are fully managed through platform APIs -- enables both high-quality data collection and holistic experimentation. Notably, the platform can now keep track of all decision points and support A/B tests between different configurations. Well-defined APIs improve adoption among product engineers with limited ML background, and ML configuration can be abstracted via declarative programming or GUI without requiring coding~\cite{molino21declarative}.

\noindent
\textbf{End-to-end AutoML.} Hyperparameter tuning is often automated per model via black-box optimization~\cite{botorch2020}.\hush{enables optimization of all aspects of the problem space.} \hush{model architecture and feature selection parameters can be optimized in a multi-objective tradeoff between model quality and compute resources~\cite{daulton2021nehvi}.}
But optimizing the loss function of SOTA models by 1\% often brings no long-term product gains, whereas tuning decision policy params usually helps, e.g., by better reflecting penalties for false positives/negatives.
In our {\em full-stack} (extended end-to-end) regime,
we enable AutoML for the entire ML pipeline via declarative {\em strategy blueprints} (Section \ref{sec:blueprints}) and an adaptive experiments framework tied to product metrics\hush{through A/B testing} \cite{bakshy2018ae}.

\subsection{Additional requirements for smart strategies}

\noindent 
\textbf{Metadata features} for product-specific models (e.g., account type, time spent online, interactions with other accounts) introduce new aspects to learning smart strategies in addition to traditional content features (images, text, video) commonly handled by ML platforms. Unlike image pixels, metadata features are diverse, uncorrelated, require non-uniform preprocessing, and are often {\em joined} from different sources. Patterns in metadata change quickly, necessitating regular retraining of ML models on fresh data, as well as monitoring and alerts. Interactions between dense metadata features can often be handled by GBDTs or shallow neural nets.
Sparse and categorical features need adequate representations~\cite{rodriguez2018beyond} and special provisions if used by neural network architectures~\cite{naumov2019dlrm}.

\hush{
\noindent
{\bf Automatic model refresh, canary and promotion} are based on daily checks of
incoming data volume: models are retrained if enough new data is collected. In addition to offline evaluations, new models are first deployed in ``canary mode". The canary model shadows predictions alongside the production model, is evaluated, and is then deployed automatically if it outperforms the production model.
}

\noindent
{\bf Non-stationary environments} are typical for deployed products but not for research prototypes and SOTA results.

\noindent
{\bf Logging and performance monitoring} are important capabilities for a production system. Dashboards monitor system health \hush{under normal circumstances} and help understand model performance in terms of statistics, distributions and trends of features and predictions, automatically triggering alerts for anomalies~\cite{amershi2019software, testscore2017}. Our platform integrates with \Meta’s online experimentation framework, and production models can be withdrawn quickly if needed.

% \noindent
% {\bf Revision control and configuration management}. As described in Section \ref{sec:blueprints}, all aspects of the strategy blueprint are saved in a version-controlled configuration storage system. Keeping these versions persisted enables robust A/B testing between different configuration versions as well as straightforward rollbacks of the entire pipeline if needed.

\noindent
{\bf Monitoring and optimizing resource usage} flags inefficiences across training and inference. Our monitoring tools track resource usage to components of the training and inference pipeline (Section \ref{sec:core}), and help trade ML performance for resources and latency. 
\hush{Less important features are found and reaped with engineers' approval (Appendix \ref{sec:resources}).}

\section{The \Looper Platform}
\label{sec:Looper}

To support a smart strategy, a vertical ML platform (Figure \ref{fig:platforms}) collects features and labels from a running product, trains a model, and produces predictions in real time for use in the product. Such "loops" need operational structure --- established processes and protocols for model revision and deployment, evaluation and tracking of product impact, and overall maintenance. 
We now introduce insights, design principles and an architecture for a vertical smart-strategies platform to address the needs outlined in Sections \ref{sec:intro} and \ref{sec:smart_strat_ml}.

% Workflows add robustness to the system and address risks,  but scalability requires a platform design.
% \hush{ The company-wide data platform from Section \ref{sec:blueprints} supports several ML platforms that vary by scope and target applications~\cite{hazelwood2018applied}. We next outline our platform philosophy and architecture, then cover key platform-level issues.} \hush{In particular, we link model tuning automation with product metrics and work with causal effects to ensure product impact.}

\subsection{Design principles and a concept inventory}
\label{sec:principles}

In contrast to heavy-weight ML models for vision, speech and NLP that favor offline inference (with batch processing) and motivate applications built around them, we address the demand for smart strategies within software applications and products. These smart strategies operate on metadata --- a mix of categorical, sparse, and dense features, often at different scales. Respective ML models are lightweight, they can be re-trained regularly and deployed quickly on shared infrastructure in large numbers. Downside risks are reduced via ($i$) simpler data stewardship, ($ii$) tracking product impact, ($iii$) failsafe mechanisms to withdraw poorly performing models. Smart strategies have a good operational safety record and easily improve naive default behaviors.

The human labeling process common for CV and NLP fails for metadata because relevant decisions and predictions (a) only make sense in an application context, (b) in cases like {\em data prefetch} (Section \ref{sec:prefetch}) only make sense to engineers, (c) may change seasonally, and even daily. Instead of human labeling, our platform interprets user-interaction and system-interaction metadata as either labels for supervised learning or rewards for reinforcement learning. \hush{Frequent retraining limits the complexity of ML models.}
\hush{The diversity of use cases served by our platform hampers transfer learning common in CV, speech and NLP applications. On the bright side, when metadata features are not coupled as tightly to each other as image pixels or elements of word embeddings, our platform selects the simplest model for the job (Section \ref{sec:smart_strat_ml}).}\hush{Online applications provide ample training data, and} %and we can afford to automatically retrain lightweight models every day on fresh data 
\hush{without augmentation or manual data prep, assuming adequate precautions}To improve operatonal safety and training efficiency, we rely on batch-mode (offline) training, even for reinforcement learning. Given real-time inference,
model agility beyond daily re-training is supported by real-time engineered features, such as event counters.

Our platform ensures fast onboarding, robust deployment and low-effort maintenance of multiple smart strategies where positive impacts are measured and optimized directly in application terms (Appendix \ref{sec:survey}).
To this end, we separate application code from platform code, and leverage existing horizontal ML platforms with interchangeable models for ML tasks (Figure \ref{fig:platforms}). Intended for company engineers, our platform benefits from high-quality data and engineered features in the company-wide feature store~\cite{featurestore2021}. To simplify onboarding for product teams and keep developers productive, we automate and support
\begin{itemize}
\item 
Workflows avoided by engineers \cite{sambasivan2021everyone}, e.g., feature selection and preprocessing, and tuning ML models for metadata.
\item 
Workflows that are difficult to reason about, e.g., tuning  ML models to product metrics. \hush{Major challenges are brought on by the need to evaluate causality effects and optimize them.}
\end{itemize}

We first introduce several concepts for platform design.

\noindent 
The \textbf{decision space} captures the shape of decisions within an application which can be made by a smart strategy. It can be just \{0,1\} to indicate whether a notification is shown. It can be a continuous-value space for time-to-live (TTL) of a cache entry. It can be a data structure with configuration values for a SW system, such as a live-video stream encoder. With reinforcement learning, the decision space matches well with the concept of action space.

\noindent 
{\bf Application context} captures necessary key information provided by a software system at inference time to make a choice in the decision space. The application context may be directly used as features or it may contain ID keys to extract the remaining features from the feature store (Section \ref{sec:blueprints}).

\noindent 
{\bf Product metrics} evaluate the performance of an application and smart strategies. When specific decisions can be judged by product metrics, one can generate labels for supervised learning, unlike for metrics that track long-term objectives.

\hush{
To illustrate these concepts on content ranking, suppose an app needs to rank a set of $N$ user-generated posts and select $M$ for display in a feed-like product surface. To configure a smart strategy, a platform user defines the input EAC to the decision API to be a set of $N$ content ids and an id for ranking recipient/viewer. The Decision space contains ordered lists of M ids. Finally the unique daily visitors of the app, and data center CPU consumption might be selected as Product metrics to be co-optimized for. 
}

%We now review what is needed to satisfy the problem spec.

\noindent 
{\bf A proxy ML task} casts product goals\hush{(in terms of product metrics)} in mathematical terms to enable ($i$) reusable ML models that optimize formal objectives and ($ii$) decision rules that map ML predictions into\hush{a form that can lead to} decisions (Section \ref{sec:blueprints_model}).
\hush{The choice of a proxy task can be accompanied by feature selection, the use of appropriate labels, additional penalty terms and re-ranking steps to improve correlation with product metrics, etc. In some cases, the appropriate proxy task is obvious and matches application needs well. In other cases, the choice between, say, supervised and reinforcement learning is unclear, and each approach requires careful modeling and/or task transformation.} 
Setting proxy tasks draws on domain expertise,
but our platofrm simplifies this process.
%, but our experience suggests room for automation.

\noindent 
{\bf Evaluation of effects on live data} verifies that solving the proxy task indeed improves product metrics. Access to \Meta's monitoring infrastructure helps detect unforeseen side effects. As in medical trials, (1) we need evidence of a positive effect, (2) side-effects should be tolerable, and (3) we should not overlook evidence of side-effects. \hush{Decoupling live evaluation of a smart strategy from the proxy ML task implies that we no longer require product developers to guarantee a strong causal relationship between action taken and outcome because we check it directly. In the past, such a relationship was ensured by hand-coding heuristics and decision rules that selected actions to drive product metrics.} On our platform, product developers define the decision space, allowing the platform to automatically select model type and hyperparameter settings. The models are trained and evaluated on live data without user impact, and improved until they can be deployed. 
Newly trained models are {\em canaried} (deployed on shadow traffic) before product use -- such models are evaluated on a sampled subset of logged features and observations, and offline quality metrics (e.g., MSE for regression tasks) are computed. This helps avoid
degrading model quality when deploying newer models.

\hush{
Continuing the ranking example, we observe that it is amenable to several types of automation, especially given the recent industry focus on feature reuse and product metrics: feature selection and feature engineering, automated label acquisition and value-model tuning, resource optimization, and management of advanced ML models. Recently the application of multi-task models has become common for high-end ranking systems \cite{zhao2019recommending}. Such models can be deployed from an ML platform like ours with support for tuning performance with respect to product metrics.
}

\hush{
Additional room for automation is found in the evaluation of causal impact on product metrics, which is commonly accomplished in the industry using A/B tests. Our interactions with platform adopters suggest not only a significant amount of routine, but also conceptual difficulties in combining multiple to product metrics into a single closed-form utility function valid across reasonable input ranges. An alternative is multi-objective optimization in terms of Pareto fronts and selection from top candidates. }

\subsection{Platform architecture: the core}
\label{sec:core}

\begin{figure}[b]
 \vspace{-2mm} 
  \centering
  \includegraphics[trim=0 0 0 30,clip,width=\linewidth]{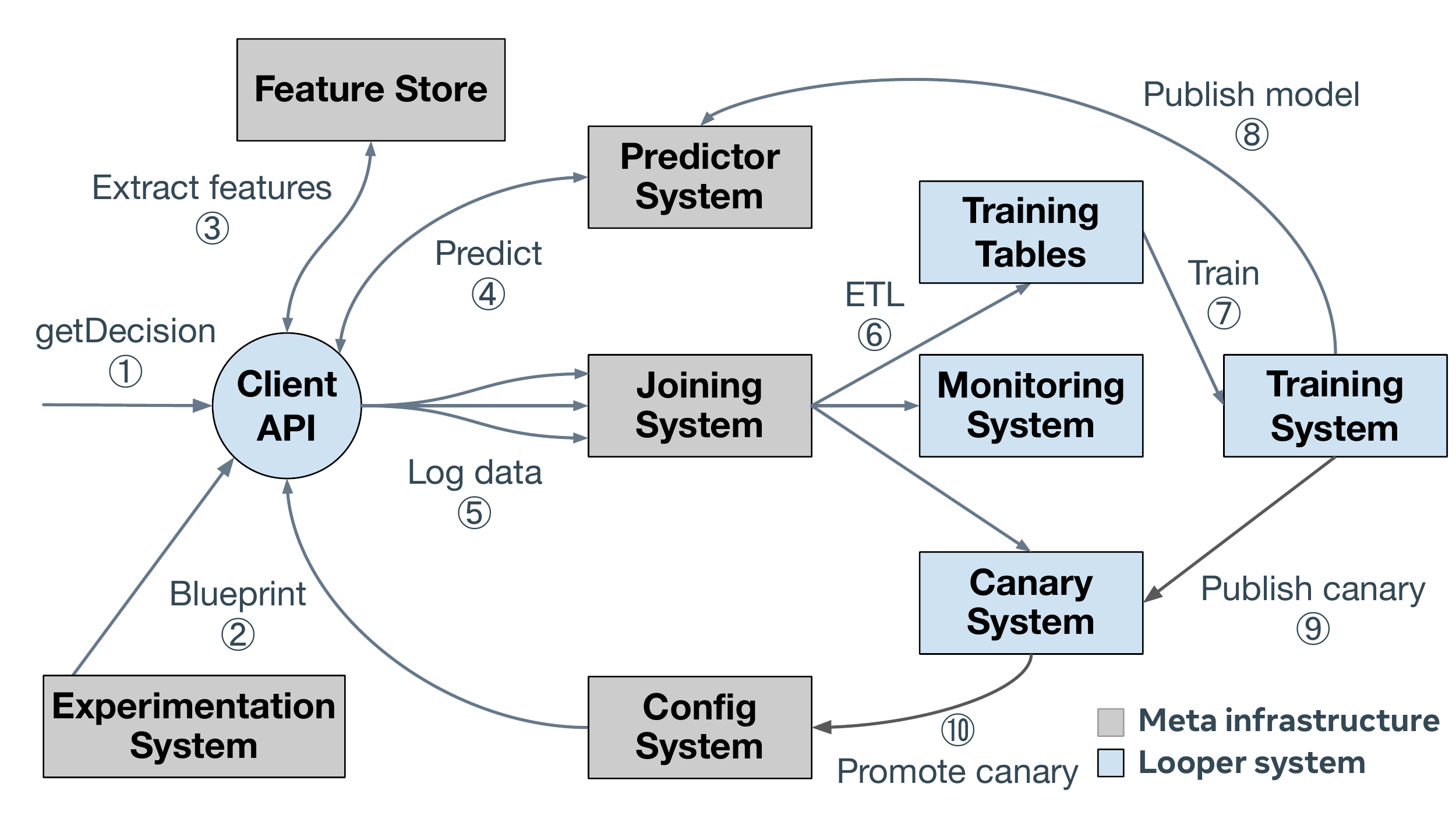}
 \vspace{-6mm} 
\caption{
  \label{fig:arch}
 Data flow in the \Looper platform.\hush{Gray boxes indicate shared \Meta infrastructure.}
 Figure \ref{fig:blueprint} expands the left side.
 }
%\vspace{-2mm}
\end{figure}

% Note: this will actually appear on the *next* page
\begin{figure*}[t]
  \centering
  \includegraphics[width=0.84\linewidth]{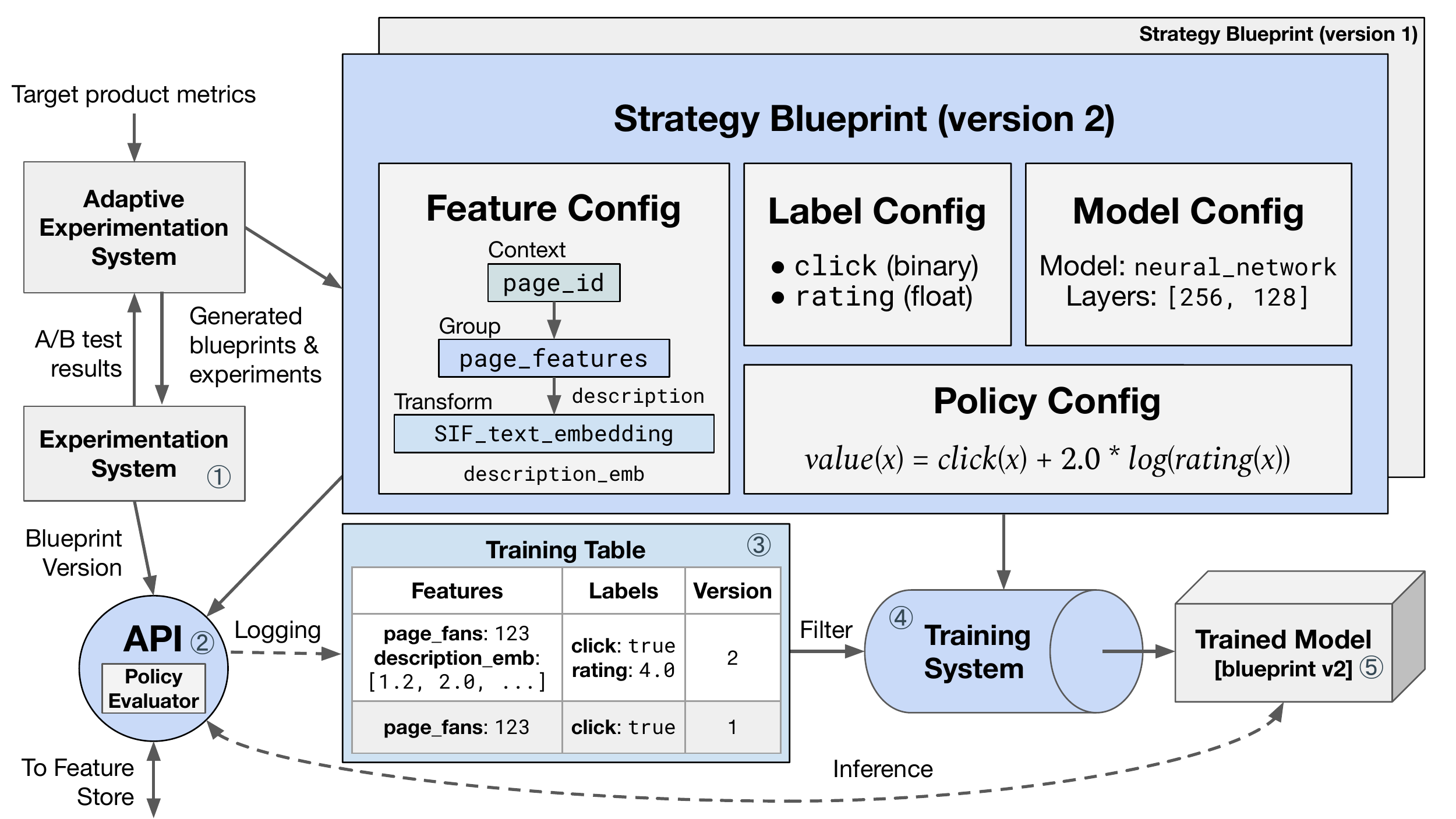}
  \vspace{-2mm}
  \caption{\label{fig:blueprint}
  The strategy blueprint and how it controls different aspects of the end-to-end model lifecycle. Continuation of Figure \ref{fig:arch}.
  }
  % https://docs.google.com/presentation/d/1QQzSL5sRkcf8j1FeRqHHL6ZTHUo2KG6cxMx5rYfmpAo/edit#slide=id.gebf84c2f45_0_0
  \vspace{-2mm}
\end{figure*}

Traditional ML pipelines build training data offline, but our platform uses a {\em live feature store} and differs in two ways:
\begin{itemize}
\item {\bf Software-centric vs. data-centric interfaces.} Rather than passed via files or databases, training data are logged from product surfaces as \Looper APIs intercept decision points in product software. Product engineers delegate training-data quality concerns (missing or delayed labels, etc) to the platform. Missing data are represented by special values.
%\vspace{-1mm}
\item {\bf An online-first approach.} \Looper API logs live features and labels at the decision and feedback points, then joins and filters them via real-time stream processing. This {\em immediate materialization} avoids data hygiene issues \cite{agarwal2016making} and storage overhead: it keeps training and inference consistent and limits label leakage by separating features and labels in time. Looper's {\em complete chain of custody for data} (without exposing data tables or files) helps prevent engineering mistakes.
\end{itemize}

\noindent
{\bf The \Looper RPC API} relies on 
two core methods:

\noindent
{\bf I.} \fbox{ $\mathtt{getDecision(decision\_id, application\_context)}$ } returns a decision-space value, e.g., $\mathtt{True/False}$ for binary choices or a floating-point score for ranking. Unlike in the 3-call APIs in \cite{agarwal2016making,carbune2018smartchoices}, $\mathtt{null}$ is returned before a model is available
(to trigger default behavior).
User-defined
$\mathtt{decision\_id}$ ties each decision with observation(s) logged later (II); it may be randomly generated for clients to use. $\mathtt{application\_context}$ is a dictionary representation of the application context (Section \ref{sec:principles}), e.g., with the user ID (used to retrieve additional user features), current date/time, etc.

\noindent
{\bf II.} \fbox{ $\mathtt{logObservations(decision\_id, observations)}$ } logs labels for training proxy ML task(s), where $\mathtt{decision\_id}$ must match a prior $\mathtt{getDecision}$ call. Observations capture users' interactions, responses to a decision (e.g., clicks or navigation actions), or environmental factors such as compute costs.

Though deceptively simple in product code, this design fully supports the MLOps needs of the platform. We separately walk through the online (inference) and offline (training) steps of the pipeline in Figure \ref{fig:arch}.
%\noindent
\circled{1}
Product code initializes the \Looper client API with one of the known strategies registered in the UI. 
$\mathtt{getDecision()}$ is then called with the $\mathtt{decision\_id}$ and $\mathtt{application\_context}$. 
\circled{2}
\Looper client API retrieves a versioned configuration (the “strategy blueprint”, Section \ref{sec:blueprints}) for the strategy to determine the features, the model instance, etc. The exact version used may be controlled through an external experimentation system.
\circled{3}
The client API passes the application context to the \Meta feature store (Section \ref{sec:blueprints}), which returns a complete feature vector.
\circled{4}
The client API passes the feature vector and production model ID to a distributed model predictor system (cf. \cite{soifer2019inference}), which returns proxy task predictions to the client.
Then, the client API uses a decision policy (Section \ref{sec:blueprints_model}) to make the final decision based on the proxy predictions. Decision policies are configured in a domain-specific language (DSL) using logic and formulas.
\circled{5}
Asynchronously, the anonymized feature vector and predictions are logged to a distributed online joining system (c.f. \cite{photon2013}), keyed by the decision ID and marked with a configurable and relatively short TTL (time-to-live). The $\mathtt{logObservations}$ API also sends (from multiple request contexts) logs to this system. Complete ``rows" with matching features and observations are logged to a training table, with retention time set per data retention policies. The remaining steps are performed offline and asynchronously.

\circled{6}
 Delayed and long-term observations are logged in a table and then joined offline via Extract, Transform, and Load (ETL) pipelines~\cite{ETLvsELT}. These pipelines perform complex data operations such as creating MDP sequences for reinforcement learning. The logged features, predictions, and observations are sent for logging and real-time monitoring as per Section \ref{sec:end2end}.
\circled{7}
An offline training system \cite{fblearner2016} retrains new models nightly or when sufficient data are available\hush{based on the training table}, addressing concerns from Section \ref{sec:principles}.
\circled{8}
Trained models are published to the distributed predictor for online inference.
\circled{9}
Models are then registered for canarying (Section \ref{sec:principles}).
\circled{10}
A canary model that outperforms the prior model is promoted to production and added to the loop configuration.

\subsection{Product optimization with strategy blueprints}
\label{sec:blueprints}

The end-to-end nature of the \Looper platform brings its own set of challenges regarding data and configuration management in the system. Existing ML management solutions~\cite{vartak2018modeldb} primarily focus on managing or versioning of data and models, which is insufficient in covering the full lifecycle of smart strategies.
In this section we introduce the concept of a \textit{strategy blueprint}, a version-controlled configuration that describes how to construct and evaluate a smart strategy. Blueprints are immutable, and modifications (typically through a GUI) create new versions that can be compared in production through an online experimentation platform, allowing for easy rollback if needed. The strategy blueprint (Figure \ref{fig:blueprint}) controls four aspects of the ML model\hush{and training} lifecycle and captures their cross-product:

\noindent \textbf{Feature configuration.} Modern ML models can use thousands of features and computed variants, which motivates a unified repository --- a \textit{feature store} --- usable in model training and real-time inference~\cite{hazelwood2018applied, featurestore2021}. Features within a \textit{feature group} are associated with the same application context (e.g., a Web page id) and are computed together. Feature variants are produced by \textit{feature transforms}, e.g., pre-trained or SIF~\cite{arora2017sif} text embeddings. The \Looper blueprint leverages\hush{the concept of} feature stores for feature management and tracks ($i$) feature groups via a computational graph and ($ii$) downstream feature transforms. Blueprint modifications often try to improve model quality by experimenting with new features.

\noindent \textbf{Label configuration} controls how ML objectives are proxied by clicks, ratings, etc, as per Section \ref{sec:blueprints_model}. Faithful proxies for product metrics are hard to find~\cite{stein2019}, hence experimentation with label sets.

\hush{
\begin{figure}[t]
%\vspace{-2mm}
  \centering
  \includegraphics[width=0.9\linewidth]{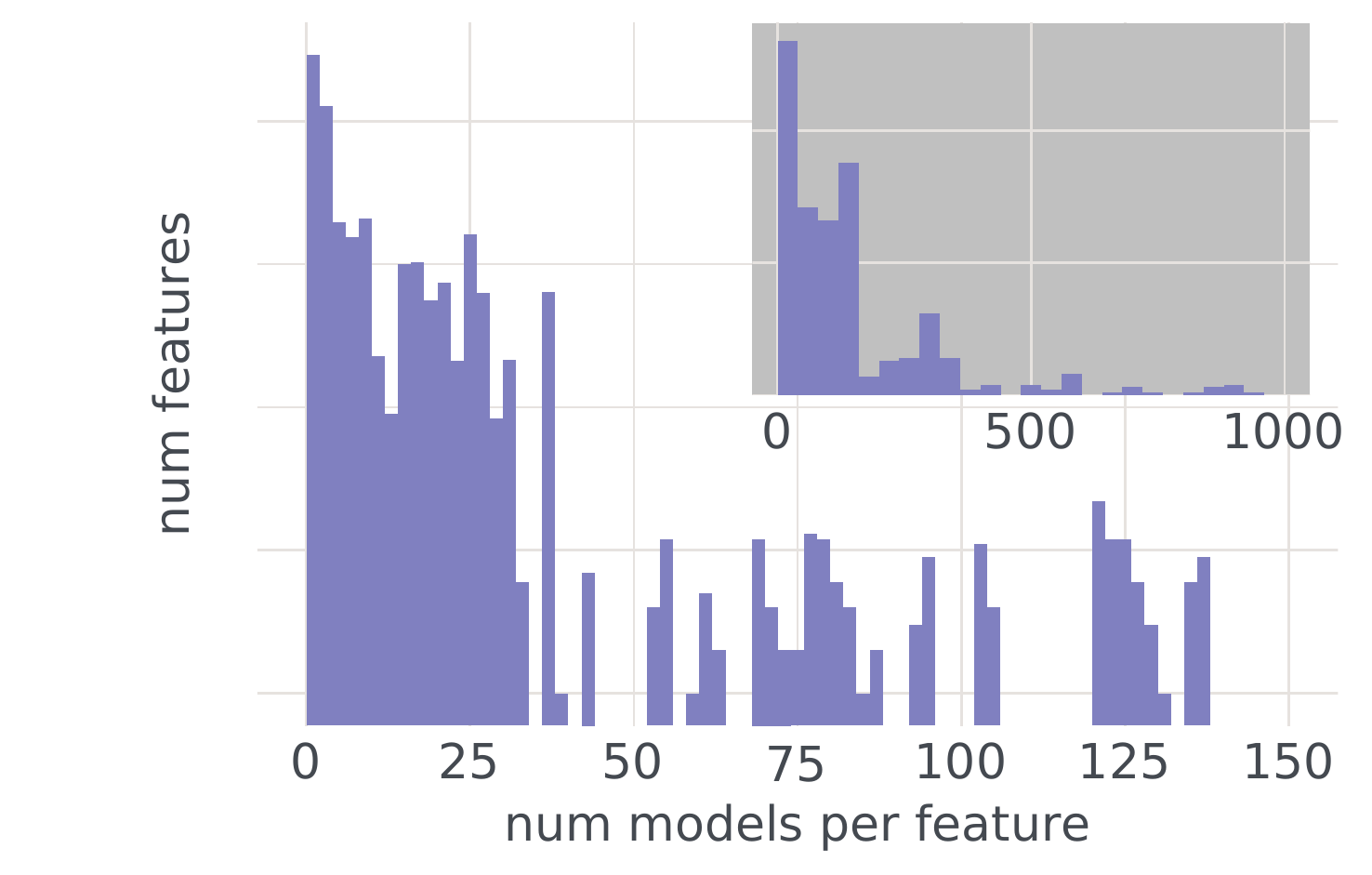}
 \vspace{-2mm} 
\caption{
\label{fig:features}
 \textbf{Outer:} A histogram of feature reuse across models. The number of features ($y$) is plotted against the reuse count ($x$). \textbf{Inner:} A histogram of feature counts per model. The number of models ($y$) is plotted against the number of features used by a model ($x$).}
\vspace{-6mm}
% https://www.internalfb.com/intern/anp/view/?id=562915
\end{figure}
}

\noindent \textbf{Model configuration} helps product teams explore model architecture tradeoffs (DNNs, GBDTs, reinforcement learning). The blueprint only specifies high-level architecture parameters, while lower-level hyperparameters (e.g., learning rates) are delegated to AutoML techniques invoked by the training system (Section \ref{sec:end2end}).

\noindent \textbf{Policy configuration} controls
how raw objective predictions are translated into decisions (Section \ref{sec:blueprints_model}). It uses a lightweight domain specific language (DSL). 
Figure \ref{fig:blueprint} illustrates a ranking decision, where the click and rating objectives are weighted and combined to generate a single score per candidate. Smart strategies often need to optimize the importance weights embedded in decision policies.

Blueprints capture \textit{compatibility} between versions, e.g., the training pipeline for ver. $A$ may use data from ver. $B$ if features and labels in $A$ are subsets of those in $B$. Tagging each training row with the originating blueprint version enables data sharing between versions.

Figure \ref{fig:blueprint} illustrates the lifecycle of a blueprint. From left to right: 
%\begin{enumerate}
%\vspace{-1mm}
\circled{1} An experimentation system enables different blueprint versions to be served across the user population to facilitate A/B testing (optionally, in concert with a ``blueprint optimizer", described later below).
\circled{2}
The client API uses the blueprint feature configuration to obtain a complete feature vector from the feature store.
\circled{3}
Completed training examples are logged to training tables, tagged with the originating blueprint version.
\circled{4}
The training system filters data by compatible version and executes the pipeline per the blueprint’s feature, label, and model configurations. The policy configuration may be needed as well for more sophisticated model types (reinforcement learning).
\circled{5}
Trained models are published in the blueprint version. So, the client API uses only models explicitly linked to its served blueprint version. To generate the final product-facing decision, the client also uses the policy configuration.
%\end{enumerate}

\noindent
{\bf Vertical optimizations} with the blueprint abstraction capture dependencies between the four configuration types. Since long-term product metrics are rarely available in closed form, such an optimization requires ($i$) a sequence of A/B tests that evaluate product metrics and ($ii$) parameter adjustment between these tests. Given that each A/B test (experiment) can take significant time and impact many end-users, very few A/B tests can be used in practice. This calls for (multi-objective) Bayesian optimization: to tune parameters in a blueprint and optimize product metrics w/o closed-form representation,
\Looper leverages an \textit{adaptive experimentation} platform~\cite{bakshy2018ae}, see
Section \ref{sec:pex_impact} for details.
Product outcomes often improve even by just tuning weights in the policy configuration, e.g., for recommendation scores and reward shaping.

\noindent
{\bf Bookkeeping} for product groups using Looper is performed in terms of {\em use cases} tied to an {\em application context} and an {\em ML task} (Section \ref{sec:principles}). Multiple candidate configurations are maintained to train and evaluate candidate {\em model instances} and define {\em decision policies}. While only one blueprint (and model) per use case is in production\eat{at any given time}, many model instances may be undergoing shadow evaluation.

\subsection{Platform specializations}
\label{sec:special}

The core platform (Sections \ref{sec:core} and \ref{sec:blueprints}) goes a long way to address the challenges listed in the Introduction.
However, additional structures are needed because
($i$) product-metric evaluation and optimization have serious blind spots, while ($ii$) some classes of application are cumbersome to support. 
Platform specializations that address these deficiencies add significant value to the platform.%

\hush{The core platform (Section \ref{sec:core}) was originally designed to be general enough to cover use cases of smart strategies. However, the growing diversity of use cases prompted the development of platform specializations for specific domains.} \hush{Here we discuss some examples of meta-platforms and how they utilize the \Looper platform as the basic building block.}

% The growing diversity of use cases prompted several specializations.

\noindent
{\bf Integrated experiment optimizations.} 
%Modern applications measure 
Even when a product metric is approximated well by an ML loss function, the correlations captured by the model might not lead to causal product improvements. Hence, A/B testing estimates the {\em average treatment effect} (ATE) of the change across product users. Shared repositories of product metrics are common
~\cite{kohavi2009controlled,bakshy2014designing,xu2015infrastructure}, and product variants are systematically explored by
running many concurrent experiments ~\cite{bakshy2018ae}.\hush{Analytical complications include the Simpson’s paradox and spurious statistical significance during multiple testing~\cite{xu2015infrastructure}.}
%Automated adaptive online experiments can fuel data-driven model selection and hyperparameter optimization \cite{dai2020model}.
% Don't think that the above adds much, also maybe a bit too much hype
While dealing with non-stationary measurements, balancing competing objectives, and supporting the design of sequential experiments \cite{bakshy2018ae}, a common challenge with A/B tests is to find subpopulations where treatment effects differ from the global ATE -- {\em heterogeneous treatment effects} (HTE). \hush{In an example unrelated to web applications, appropriate medication dosage may grow with body weight, and understanding this link is critical to pharmaceutical studies. The dosage that works best on average may be ineffective or even dangerous to some individuals.}
Common neglect for HTEs in A/B testing leaves room for improvement~\cite{bakshy2014designing, beutel2017beyond}\hush{in political science, medicine, and technology}~\cite{wager2018estimation}, likely delivering suboptimal treatments. The \Looper platform and its support for A/B testing dramatically simplify HTE modeling on the \Meta online experimentation platform, and help deploying treatment assignments based on HTE estimates.

In an initial training phase, \Looper's $\mathtt{getDecision()}$ API acts as a drop-in replacement for the standard A/B testing API, and falls through to a standard randomized assignment while still logging features for each experiment parti\-cipant. Then, metrics from the standard A/B testing repertoire help derive the treatment outcome (observations) for each parti\-cipant, and the \Looper platform trains specialized HTE models (meta-learners such as T-, X-, and S- learners \cite{kunzel2019meta}). In a final step, the HTE model predictions can be used in a decision policy to help $\mathtt{getDecision()}$ make intelligent treatment assignments and measurably improve outcomes compared to any individual treatment alone. In this scenario, the best HTE estimate for a given user selects the actual treatment group.
Our integration links \Looper to an established experiment optimization system \cite{bakshy2018ae} and creates synergies discussed in Section \ref{sec:pex_impact}.
A further extension relaxes the standard A/B testing contract to support {\em fully dynamic assignments} and enables reinforcement learning\hush{for assignment optimization}\anoncite{apostolopoulos2021personalization}.

\noindent
{\bf \Looper for ranking.}
The $\mathtt{getDecision()}$ + $\mathtt{logObservations()}$ API is general enough to implement simple recommendation systems, but advanced systems need finer support. Higher-ranked items are more often chosen by users, and this {\em positional bias} can be handled (in the API) by including the displayed position as a special input during training~\cite{craswell2008experimental}. To derive a final priority score for each item, the multiple proxy task predictions are often combined through a weighted combination function~\cite{zhao2019recommending}. Recommender systems learn from user feedback, as long ass lesser-explored items are occassionally included among top results (the explore/exploit tradeoff~\cite{yankov2015evaluation}). A specialized \Looper ranking system abstracts these considerations in a higher-level API ($\mathtt{getRanking}$) allowing the ordering of an entire list of application contexts, and also allows recording of display-time observations such as the relative screen position of each item.

\section{Production deployment}
\label{sec:impact}
\hush{
Some of the most visible applications of ML to large software systems are built around large ML models for recommendation systems, speech processing, image and video classification, etc. These models require great computational resources and dedicated infrastructure teams.}

Looper supports real-time inference with moderate-sized models to improve various aspects of software systems. 
These models are trained and deployed quickly and maintained on the platform, whereas our two-call RPC API (Section \ref{sec:core}) decouples platform code from application code.
% We have previously developed optimized prefetch strategies using RNNs~\cite{wang2019predictive} and a RL-based personalization for Web-based services~\cite{apostolopoulos2021personalization}. 
Deployed in production at \Meta during the entire year 2021, \Looper improved product metrics, made use of compute resources more efficient, and streamlined maintenance. To enhance the reproducibility of our work, 
we describe empirical observations and prominent applications, outline adoption and impact,
and summarize adopter survey results (in
Appendix~\ref{sec:survey}). Resource estimates are in "server" units (we do not use GPUs). For smart strategies that learn from end-users, performance and impact inevitably depend on the user base and product metrics infrastructure. Hence, we report \Looper impact as a fraction of product teams's half-year results. Resource savings are reported as percent vs. baselines.%

%We illustrate the diversity of applications by two %types -- causal product-metric optimization and data %prefetching. Then we summarize the overall product %adoption and impact of \Looper.

\begin{figure}[t]
\vspace{-2mm}
  \centering
  \includegraphics[width=0.8\linewidth]{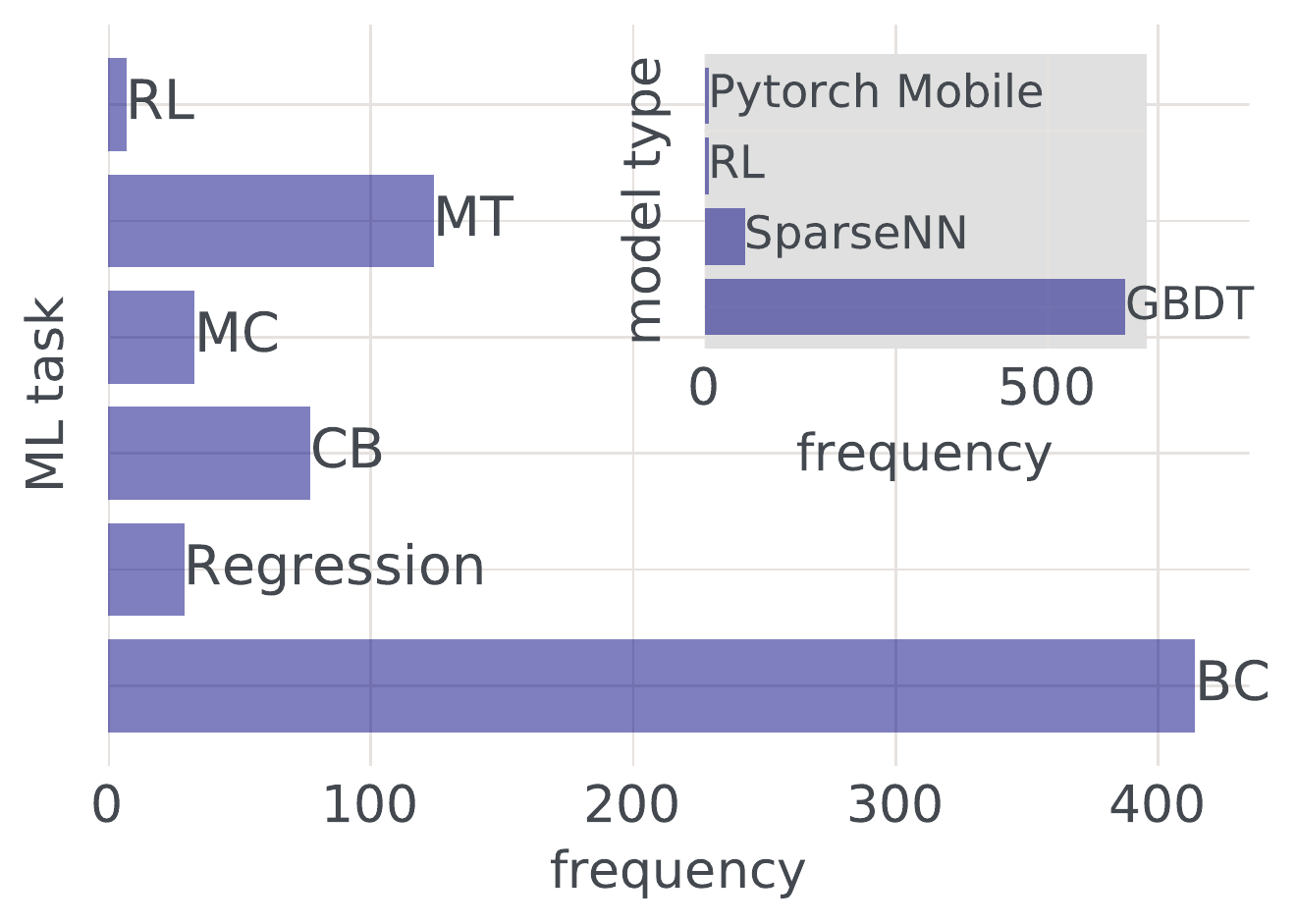}
\vspace{-4mm}
\caption{
  \label{fig:models}
ML tasks and model types on our platform. On the outer plot --- ML tasks: Reinforcement Learning (RL), multitask (MT), multiclass (MC) and binary  classification (BC), contextual bandit (CB), regression. Multitask models blend \hush{several binary} classification or regression sub-tasks. On the inner plot --- model types: 
PyTorch Mobile, RL, NN, Gradient Boosted Decision Trees. }
% https://www.internalfb.com/intern/anp/view/?id=562915
\vspace{-2mm}
\end{figure}

\subsection{Statistics for ML tasks and models in \Looper}
\label{sec:numbers}
Given nonstationary application environments,
\Looper retrains models on the day when 20\% fresh training data become available (or on a set schedule). Choosing appropriate ML models requires trading off performance metrics with resource usage, inference latency, and configuration effort. SVM packages struggle with multimodal data and scale poorly to voluminous data. DNNs scale to 1B+ data rows, handle sparse features, but use more memory than simpler models.
DNNs are sensitive to architecture configuration and are currently less explored for tabular data, making configuration challenging without ML expertise.
Gradient-Boosted Decision Trees (GBDT, XGBoost) are compact and robust, handle multimodal tabular data naturally (including sparse features and missing values), do not require architecture search or GPUs, and scale to 100M rows. \hush{They are often {\em good enough} and use moderate resources.}
Typical {\em inference latency} with GBDT and XGBoost is in low single $ms$ for server-side models and 1-2 ms for (smaller) mobile models.
Inference with DNN-based models, especially those using latent embeddings, is an order of magnitude slower. That's why the mean inference latency (10ms) across Looper use cases is greater than the median (2ms). 90\%le and 99\%le latencies are 20ms. Figure \ref{fig:models} summarizes ML tasks deployed on \Looper and the models selected for them. 
Figure \ref{fig:features} shows that models typically use 50-200 features, and most features are used by many models. Feature extraction latency has median 45ms and mean 120ms. Latencies for extr†acting synthetic/engineered features are greater (90\%le and 99\%le is 240 ms).
% Additionally, in our experience (Figure \ref{fig:models}), CB and RL models often do not yield sufficient gains to justify the additional complexity.
%
For a broader picture, Figure \ref{fig:use_case_resource_type} summarizes 29 utilization rates for system resources. In particular, feature extraction tends to be a bottleneck. This data is further discussed in Appendix \ref{sec:resources} and used to optimize resource utilization in \Looper.%

\subsection{Application deep dive -- personalized experiments}
\label{sec:pex_impact}

  Section \ref{sec:special}, while focusing on our platform architecture, outlined a special application of smart strategies ---
  embedding them into the standard experimentation framework to automate the personalization of A/B treatments when HTEs are detected and captured by ML models. As A/B testing APIs are common and accessible~\cite{bakshy2014designing, xu2015infrastructure}, \Looper's integration with the standard experimentation APIs makes 
  training and deploying personalized smart strategies as easy as changing \textit{one or two lines} of code by product teams.
%  Given that many companies developed A/B testing APIs
  In practice, exposing smart strategies through such APIs
  brings several benefits:
\begin{itemize}
\item 
Client code and the learning curve are simplified
by repurposing the decision API as the A/B testing API.

\item Dataset preparation and modeling flow are automated for the task of optimizing metric responses based on end-users exposed to each treatment. Metric responses can be automatically sourced from the experimentation measurement framework without manual labeling.

\item 
Product metrics are traded off across many strategies, offline and online, via multi-objective optimization (MOO) \cite{bakshy2018ae,daulton2021nehvi}.

\item
Smart strategies are automatically compared against all baseline treatments (i.e., A or B), making the tradeoff between metric impact and costs explicit.

\end{itemize}
Previously such experiment optimization needed dedicated engineering resources. Now the tight integration of the \Looper platform with the experimentation framework allows product engineers quickly evaluate a smart strategy and optimize its product impact {\em in several weeks}. With automatic MOO, engineers find tradeoffs appropriate to a given product context. For example, during a server capacity crunch, one team traded a slight deterioration in a product metric for a 50\% resource savings.\eat{Per month, three to four adaptive product experiments launched via \Looper use integrated experiment optimization for smart A/B testing and parameter optimization.} Predicating product deployment on such experiments creates safeguards
against ML models that generalize poorly to live data. This also helps tracking product impact.
For example, a user authentication use case
\cite{apostolopoulos2021personalization}
reduced SMS cost by 5\% while remaining neutral for engagement metrics.

\hush{
\begin{figure}[t]
  \centering
  \includegraphics[width=0.99\linewidth]{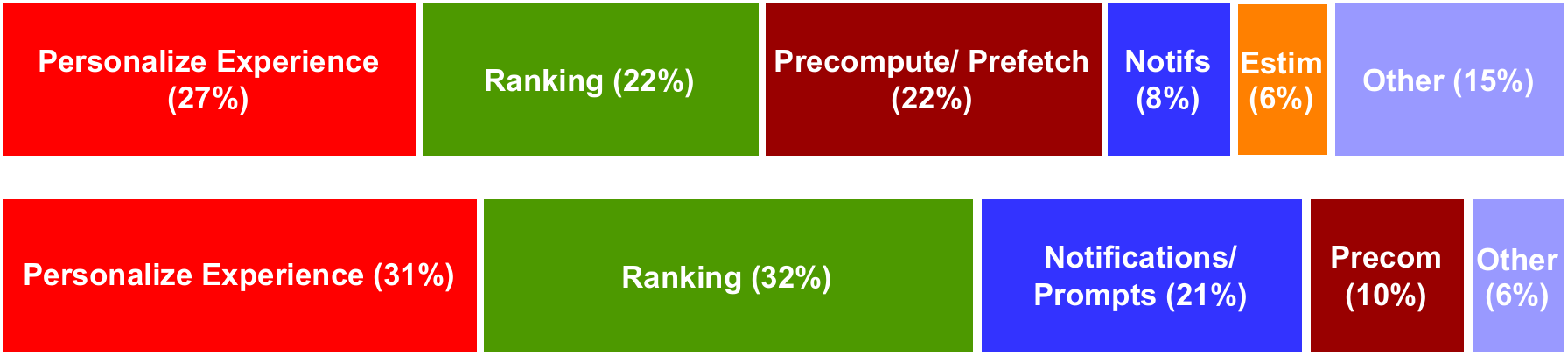}
  \vspace{-1mm}                                     
  \caption{\label{fig:use_case_type} Product adoption of smart strategies by use case (top) and by resource consumption (bottom)
  }
  \vspace{-3mm}                             
\end{figure}
}

%  \noindent
\hush{and investments into hardware accelerators amortized over numerous applications.} \hush{In particular, we developed a prioritization methodology for new optimizations and upgrades. For example, when updating model calibration, we aggregated application impacts weighed by estimates of relative importance based on total historical utilization volume with exponential decay over time. Importance can be adjusted based on business context. Having a single figure of merit makes go/no-go decisions more consistent and transparent. Software and hardware optimizations unaffordable for a single application become practical for an ML platform.}
\hush{Marketplace has a dedicated infrastructure team responsible for interfacing with the company’s central capacity-planning function to get sufficient computational resources to support the organization’s needs, including the ML use cases. The FBLite team had to request quota increases for real time logging and counter services as they scaled the number of use cases on the Smart Strategies Platform.}

\subsection{Application deep dive -- data prefetching}
\label{sec:prefetch}

%We now illustrate applications that adopted smart strategies. 
\hush{
On edge devices, prefetching sections of our company's large software made a
significant impact. This technique makes decisions based on historical user activity. We also introduce emerging applications of federated learning to support privacy-aware smart strategies.}

Online applications strive to reduce 
response latency for user interactions.
Optimized resource prefetching based on user history helps by proactively loading application data. Modern ML methods can accurately predict the likelihood of data usage, minimizing unused prefetches.
Our \Looper platform supports prefetching strategies for many systems within \Meta, often deeply integrated into the product infrastructure stack. For example, \Meta's GraphQL\anoncite{graphql2015} data fetching subsystem uses our platform to decide which prefetch requests to service, saving both client bandwidth and server-side resources. It yields around 2\% compute savings at peak server load.
As another example, \Meta's application client for lower-end devices (with a “thin-client” server-side rendering architecture\anoncite{fblite2016}) also uses our platform to predictively render entire application screens. Our automated end-to-end system helps deploying both models and threshold-based decision policies then tune them for individual GraphQL queries or application screens, with minimal engineering effort.
Based on numerous deployed prefetch models, we have also developed large-scale modeling of prefetching. User-history models have proven to be helpful for this task\anoncite{wang2019predictive}; building up on this idea, we created application-independent vector embeddings based on users’ surface-level activity.\hush{across all \Meta surfaces.}
To accomplish this, we train a multi-task auto-regressive neural-network model to predict how long a user will stay in each application surface (e.g., news feed, search, notifications), based on a sequence of {\tt (application surface, duration)} events from the user’s history. As is common in CV and NLP, intermediate-layer outputs of this DNN predict prefetch accesses well and make specialized features unnecessary.
Optimized prefetching illustrates how secondary, domain-specific platforms are enabled by the core \Looper platform; infrastructure teams only need to wire up the prediction and labeling integration points while \Looper provides full ML support.

\subsection{Adoption and impact}
\label{sec:adoption_impact}

\begin{figure}[t]
%\vspace{-1mm}
  \centering
  \includegraphics[width=0.99\linewidth]{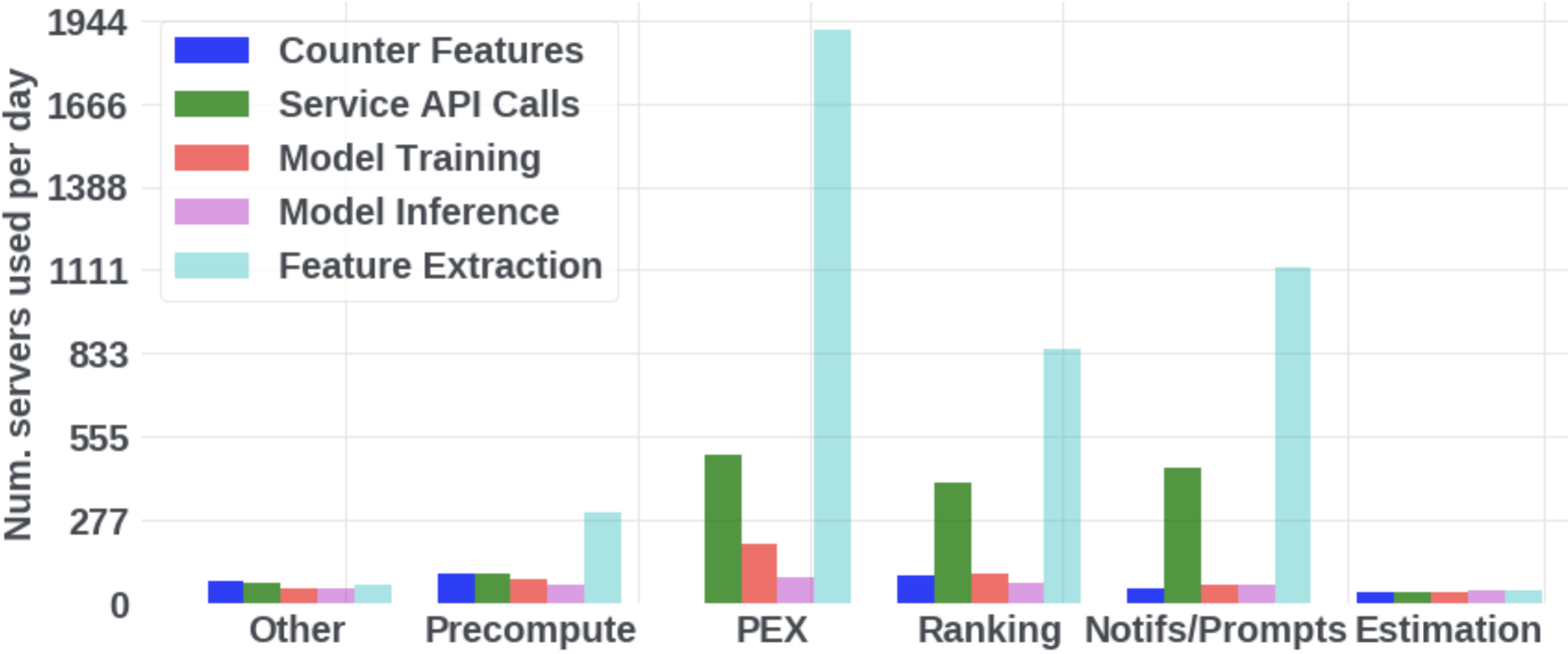}
 % \includegraphics[width=0.99\linewidth]{use_case_resource_type_usage_new.pdf}
%\
\vspace{-3mm}
  \caption{\label{fig:use_case_resource_type} Resource utilization rate\eat{(servers)} by resource type and application category (see Section \ref{sec:adoption_impact}). The Service API category includes API calls other than feature extraction and prediction service.
  }
\vspace{-4mm}
\end{figure}

 Several internal vertical platforms at \Meta\anoncite{hazelwood2018applied} compete for a rich and diverse set of applications.  Product teams sometimes relocate their ML models to a platform with greater advantages, while a few high-value applications are run by dedicated infrastructure teams. \Looper was chosen and is currently used by 90+ product teams at \Meta.\hush{, 49\% of which are in product pillars, including 31\% under Growth and 15\% under Infrastructure.}
 On any day in 2021, these teams deployed 440-1K models\hush{in 250+ use cases} that made 4-6 million decisions per second.
 Application use cases fall into five categories in decreasing order of usage (Figures \ref{fig:use_case_resource_type} and \ref{fig:use_case_type}):
 \begin{itemize}
\item {\bf Personalized Experience} is tailored based on the user's engagement history. For example, we display a new feature prominently only to those likely to use it.
\item {\bf Ranking} orders items to improve user utility, e.g., to personalize a feed of candidate items for the viewer.
\item {\bf Prefetching/precomputing} data/resources based on predicted likelihood of usage (Section \ref{sec:prefetch}).
\item {\bf Notifications/prompts} can be gated on a per-user basis, and sent only to users who find them helpful.
% For example, Marketplace only badges the Marketplace tab if a user is likely to engage with the tab.
\item {\bf Value estimation} predicts regression tasks, e.g., latency or memory usage of a data query.
 \end{itemize}

The impact of ML performance on product metrics varies by application. For a binary classifier, increasing ROC AUC from 90\% to 95\% might not yield large product gains when such decisions contribute little to product metrics if bottlenecks lie elsewhere. But increasing ROC AUC from 55\% to 60\% is impactful when each percent translates into tangible resource savings or other metrics, as it would be for online payment processing.
%
% By optimizing product impact through causal effects, \Looper use cases have made significant impacts to end metrics at \Meta including: Monthly Active People, Meaningful interactions, App Fluidity, Peak Data Center Usage, and Revenue.
%
% We regularly onboard new use cases, e.g., (a) the Marketplace-wide badging system uses \Looper as the core modeling component, (b) a search tool was built for security investigators to sift through a large corpus when responding to incidents.\hush{Our platform's impact can also be observed}
%
\Looper use cases contributed to compute savings (server utilization), user engagement (e.g., daily active users) and
other top-line company reporting metrics. Many product teams at \Facebook and \Instagram adopted \Looper without additional staffing, and it is common for \Looper to contribute 20-40\% of improvements to product goal metrics. In several cases, \Looper helped product teams outperform their goals by over 2x.
% often exceeding six-month goals.

\section{Conclusions}
\label{sec:conclusions}
We outline opportunities to embed data-driven self-optimizing smart strategies for product decisions into software systems, so as to enhance user experience, optimize resource utilization, and support new functionalities. We describe the deployment of smart strategies (at an unprecedented scale) through software-centric ML integration where decision points are intercepted and data is collected through APIs \cite{agarwal2016making}. \hush{while choosing relevant ML techniques supports application impact.} This process requires infrastructure and automation\hush{helps select ML models and configure them for each task to reduce resource needs. Given the limited supply of ML engineers,} to reduce operational mistakes and maintain ML development velocity.

% Unfortunately, infrastructure and engineering costs of modern ML models, their development, and aggressive optimization can be exorbitant to all but the largest companies and teams, so practical success requires balancing such costs against ML model performance as well as actual product impact. Another way of looking into these tradeoffs is to formulate and optimize an overall utility function.

%\noindent
Our ML platform \Looper addresses the complexities of product-driven end-to-end ML systems and facilitates at-scale deployment of smart strategies through technical insights, 
platform-level abstractions, a novel architecture, the use of Bayesian optimization, and interfaces with an adaptive experimentation system.\hush{As a platform, \Looper incorporates (1) software and reusable resources that can be configured and quickly adapted to different problems or products, (2) workflows that use these assets in various scenarios.}\hush{Successful platforms support entire ecosystems of services and users around them that increase reuse, leverage experiences, warn about pitfalls, and offer attractive job opportunities to experienced engineers. Compared to generic ML-centric software platforms such as PyTorch \cite{pytorch2020}, TensorFlow \cite{tensorflow2016}, etc, we promote system platforms with links to live data and product metrics, offer model hosting capabilities, track and optimize resource utilization, while also paying attention to engineering costs. Some of these differences are illustrated by Tensor Flow Extended (tFX) - an end-to-end platform for deploying production ML pipelines.} 
As an important simplification,
inference input processing matches
that for training. The \Looper product RPC API is simplified down to two calls. The \Looper platform treats end-to-end ML development more broadly than prior work \cite{molino21declarative, wu2021sustainable}, providing extensive support for product impact evaluation of smart strategies via causal inference.\hush{It offers a broad range of ML techniques and extensive support for mobile platforms.} \Looper learns heterogenous treatment effects (HTE) from product evaluation data by repurposing the \Looper RPC API as a drop-in replacement for a standard A/B testing API. 
Vertical optimizations with long-term product objectives are enabled by our {\em strategy blueprint abstraction} and the use of Bayesian optimization.

As observed during production deployment in 2021, \Looper offers immediate, tangible benefits in terms of data availability, easy configuration, judicious use of available resources, reduced engineering effort, and ensuring product impact. It makes smart strategies easily accessible to product engineers\hush{as in \cite{agarwal2016making,carbune2018smartchoices}} 
at large scale and enables product teams to build, deploy and improve ML-driven capabilities in a self-serve fashion without ML expertise. We observed product teams launch smart strategies within their products in one month 
(Appendix \ref{sec:survey}). The lower barriers to entry and faster deployment lead to more pervasive use of ML to optimize user experience in new products and old products not designed with ML in mind.
 \hush{In this paper, we pay particular attention to most promising application categories, the adoption process, and common roadblocks.}
Support for prefetching and personalized A/B testing have been in demand, whereas end-to-end management enables holistic resource accounting and optimization \cite{wu2021sustainable}.
The overall impact in product metrics and resource-efficiency is substantial. We also provide empirical insights into usage by resource types and application category.

Long-term benefits of our platform approach include effort and module reuse, end-to-end reproducibility, consistent reporting, reliable maintenance, and being able to upgrade ML libraries and offer new ML model types with consistent interface. Successful \Looper adopters often launch additional data-driven smart strategies, and this virtuous cycle encourages designing SW systems with built-in ML to enhance user experience and adaptation to the environment.
\hush{An ecosystem with many active long-term platform adopters can generate network-effect synergies, such as the sharing of experience, help with troubleshooting, on-demand engineering resources, etc. Our platform \Looper shares some commonalities with Overton from Apple, described by the authors as “a system to help engineers manage the lifecycle of production machine learning systems.” These commonalities include the support of code-free instantiation of ML models via declarative programming and higher-level abstractions. Furthermore, \Looper’s main purpose is to facilitate the embedding of smart strategies into modern software systems, even if ML capabilities were not considered in system design.}
 
% We discussed them in Sections~\ref{sec:intro} and~\ref{sec:survey}, and described the impact of \Looper adoption with detailed statistics in Section~\ref{sec:impact}.}


%\vspace{-2mm}
\bibliographystyle{ACM-Reference-Format}
\bibliography{Smart}


\begin{thebibliography}{55}

%%% the Web address.

\ifx \showCODEN    \undefined \def \showCODEN     #1{\unskip}     \fi
\ifx \showDOI      \undefined \def \showDOI       #1{#1}\fi
\ifx \showISBNx    \undefined \def \showISBNx     #1{\unskip}     \fi
\ifx \showISBNxiii \undefined \def \showISBNxiii  #1{\unskip}     \fi
\ifx \showISSN     \undefined \def \showISSN      #1{\unskip}     \fi
\ifx \showLCCN     \undefined \def \showLCCN      #1{\unskip}     \fi
\ifx \shownote     \undefined \def \shownote      #1{#1}          \fi
\ifx \showarticletitle \undefined \def \showarticletitle #1{#1}   \fi
\ifx \showURL      \undefined \def \showURL       {\relax}        \fi
% invisible to TeX
\providecommand\bibfield[2]{#2}
\providecommand\bibinfo[2]{#2}
\providecommand\natexlab[1]{#1}
\providecommand\showeprint[2][]{arXiv:#2}

\bibitem[\protect\citeauthoryear{Abadi et~al\mbox{.}}{Abadi
  et~al\mbox{.}}{2016}]%
        {tensorflow2016}
\bibfield{author}{\bibinfo{person}{Martín Abadi} {et~al\mbox{.}}}
  \bibinfo{year}{2016}\natexlab{}.
\newblock \showarticletitle{TensorFlow: A System for Large-Scale Machine
  Learning}. In \bibinfo{booktitle}{\emph{OSDI}}. \bibinfo{pages}{265--283}.
\newblock


\bibitem[\protect\citeauthoryear{Agarwal et~al\mbox{.}}{Agarwal
  et~al\mbox{.}}{2016}]%
        {agarwal2016making}
\bibfield{author}{\bibinfo{person}{Alekh Agarwal} {et~al\mbox{.}}}
  \bibinfo{year}{2016}\natexlab{}.
\newblock \showarticletitle{Making contextual decisions with low technical
  debt}.
\newblock \bibinfo{journal}{\emph{arXiv:1606.03966}} (\bibinfo{year}{2016}).
\newblock


\bibitem[\protect\citeauthoryear{Agarwal, Chen, and Elango}{Agarwal
  et~al\mbox{.}}{2009}]%
        {agarwal2009explore}
\bibfield{author}{\bibinfo{person}{Deepak Agarwal}, \bibinfo{person}{Bee-Chung
  Chen}, {and} \bibinfo{person}{Pradheep Elango}.}
  \bibinfo{year}{2009}\natexlab{}.
\newblock \showarticletitle{Explore/exploit schemes for web content
  optimization}. In \bibinfo{booktitle}{\emph{ICDM 2009}}.
  \bibinfo{pages}{1--10}.
\newblock


\bibitem[\protect\citeauthoryear{Amershi et~al\mbox{.}}{Amershi
  et~al\mbox{.}}{2019}]%
        {amershi2019software}
\bibfield{author}{\bibinfo{person}{Saleema Amershi} {et~al\mbox{.}}}
  \bibinfo{year}{2019}\natexlab{}.
\newblock \showarticletitle{Software engineering for machine learning: A case
  study}. In \bibinfo{booktitle}{\emph{ICSE-SEIP}}. \bibinfo{pages}{291--300}.
\newblock


\bibitem[\protect\citeauthoryear{Ananthanarayanan
  et~al\mbox{.}}{Ananthanarayanan et~al\mbox{.}}{2013}]%
        {photon2013}
\bibfield{author}{\bibinfo{person}{Rajagopal Ananthanarayanan} {et~al\mbox{.}}}
  \bibinfo{year}{2013}\natexlab{}.
\newblock \showarticletitle{Photon: Fault-Tolerant and Scalable Joining of
  Continuous Data Streams}. In \bibinfo{booktitle}{\emph{ACM SIGMOD Int'l Conf.
  on Management of Data}} \emph{(\bibinfo{series}{SIGMOD '13})}.
  \bibinfo{publisher}{ACM}, \bibinfo{pages}{577–588}.
\newblock
\showISBNx{9781450320375}
\urldef\tempurl%
\url{https://doi.org/10.1145/2463676.2465272}
\showDOI{\tempurl}


\bibitem[\protect\citeauthoryear{Anonymous}{Anonymous}{2021}]%
        {ETLvsELT}
\bibfield{author}{\bibinfo{person}{Anonymous}.}
  \bibinfo{year}{2021}\natexlab{}.
\newblock \bibinfo{title}{ETL vs ELT: Must Know Differences}.
\newblock
\newblock
\urldef\tempurl%
\url{https://www.guru99.com/etl-vs-elt.html}
\showURL{%
\tempurl}


\bibitem[\protect\citeauthoryear{Apostolopoulos et~al\mbox{.}}{Apostolopoulos
  et~al\mbox{.}}{2021}]%
        {apostolopoulos2021personalization}
\bibfield{author}{\bibinfo{person}{Pavlos~Athanasios Apostolopoulos}
  {et~al\mbox{.}}} \bibinfo{year}{2021}\natexlab{}.
\newblock \showarticletitle{Personalization for Web-based Services using
  Offline Reinforcement Learning}.
\newblock \bibinfo{journal}{\emph{arXiv:2102.05612}} (\bibinfo{year}{2021}).
\newblock


\bibitem[\protect\citeauthoryear{Arora, Liang, and Ma}{Arora
  et~al\mbox{.}}{2017}]%
        {arora2017sif}
\bibfield{author}{\bibinfo{person}{Sanjeev Arora}, \bibinfo{person}{Yingyu
  Liang}, {and} \bibinfo{person}{Tengyu Ma}.} \bibinfo{year}{2017}\natexlab{}.
\newblock \showarticletitle{A Simple but Tough-to-Beat Baseline for Sentence
  Embeddings}.
\newblock \bibinfo{journal}{\emph{ICLR}} (\bibinfo{year}{2017}).
\newblock


\bibitem[\protect\citeauthoryear{Bakshy et~al\mbox{.}}{Bakshy
  et~al\mbox{.}}{2018}]%
        {bakshy2018ae}
\bibfield{author}{\bibinfo{person}{Eytan Bakshy} {et~al\mbox{.}}}
  \bibinfo{year}{2018}\natexlab{}.
\newblock \showarticletitle{{AE}: {A} domain-agnostic platform for adaptive
  experimentation}.
\newblock \bibinfo{journal}{\emph{NeurIPS 2018 Systems for ML Workshop}}.
\newblock


\bibitem[\protect\citeauthoryear{Bakshy, Eckles, and Bernstein}{Bakshy
  et~al\mbox{.}}{2014}]%
        {bakshy2014designing}
\bibfield{author}{\bibinfo{person}{Eytan Bakshy}, \bibinfo{person}{Dean
  Eckles}, {and} \bibinfo{person}{Michael~S Bernstein}.}
  \bibinfo{year}{2014}\natexlab{}.
\newblock \showarticletitle{Designing and deploying online field experiments}.
  In \bibinfo{booktitle}{\emph{WWW' 14}}. \bibinfo{pages}{283--292}.
\newblock


\bibitem[\protect\citeauthoryear{Balandat, Karrer, Jiang, Daulton, Letham,
  Wilson, and Bakshy}{Balandat et~al\mbox{.}}{2020}]%
        {botorch2020}
\bibfield{author}{\bibinfo{person}{Maximilian Balandat}, \bibinfo{person}{Brian
  Karrer}, \bibinfo{person}{Daniel~R. Jiang}, \bibinfo{person}{Samuel Daulton},
  \bibinfo{person}{Benjamin Letham}, \bibinfo{person}{Andrew~Gordon Wilson},
  {and} \bibinfo{person}{Eytan Bakshy}.} \bibinfo{year}{2020}\natexlab{}.
\newblock \showarticletitle{BoTorch: A Framework for Efficient Monte-Carlo
  Bayesian Optimization}. In \bibinfo{booktitle}{\emph{NeurIPS 33}}.
\newblock


\bibitem[\protect\citeauthoryear{Beutel, Chi, Cheng, Pham, and Anderson}{Beutel
  et~al\mbox{.}}{2017}]%
        {beutel2017beyond}
\bibfield{author}{\bibinfo{person}{Alex Beutel}, \bibinfo{person}{Ed~H Chi},
  \bibinfo{person}{Zhiyuan Cheng}, \bibinfo{person}{Hubert Pham}, {and}
  \bibinfo{person}{John Anderson}.} \bibinfo{year}{2017}\natexlab{}.
\newblock \showarticletitle{Beyond globally optimal: Focused learning for
  improved recommendations}. In \bibinfo{booktitle}{\emph{WWW' 17}}.
  \bibinfo{pages}{203--212}.
\newblock


\bibitem[\protect\citeauthoryear{Breck, Cai, Nielsen, Salib, and Sculley}{Breck
  et~al\mbox{.}}{2017}]%
        {testscore2017}
\bibfield{author}{\bibinfo{person}{Eric Breck}, \bibinfo{person}{Shanqing Cai},
  \bibinfo{person}{Eric Nielsen}, \bibinfo{person}{Michael Salib}, {and}
  \bibinfo{person}{D. Sculley}.} \bibinfo{year}{2017}\natexlab{}.
\newblock \showarticletitle{The ML Test Score: A Rubric for ML Production
  Readiness and Technical Debt Reduction}. In \bibinfo{booktitle}{\emph{IEEE
  Big Data}}.
\newblock


\bibitem[\protect\citeauthoryear{Byron}{Byron}{2015}]%
        {graphql2015}
\bibfield{author}{\bibinfo{person}{Lee Byron}.}
  \bibinfo{year}{2015}\natexlab{}.
\newblock \bibinfo{title}{GraphQL: A data query language}.
\newblock
\newblock
\urldef\tempurl%
\url{https://engineering.fb.com/2015/09/14/core-data/graphql-a-data-query-language}
\showURL{%
\tempurl}


\bibitem[\protect\citeauthoryear{Carbune, Coppey, Daryin, Deselaers, Sarda, and
  Yagnik}{Carbune et~al\mbox{.}}{2018}]%
        {carbune2018smartchoices}
\bibfield{author}{\bibinfo{person}{Victor Carbune}, \bibinfo{person}{Thierry
  Coppey}, \bibinfo{person}{Alexander Daryin}, \bibinfo{person}{Thomas
  Deselaers}, \bibinfo{person}{Nikhil Sarda}, {and} \bibinfo{person}{Jay
  Yagnik}.} \bibinfo{year}{2018}\natexlab{}.
\newblock \showarticletitle{{SmartChoices}: Hybridizing programming and machine
  learning}.
\newblock \bibinfo{journal}{\emph{arXiv:1810.00619}} (\bibinfo{year}{2018}).
\newblock


\bibitem[\protect\citeauthoryear{Covington, Adams, and Sargin}{Covington
  et~al\mbox{.}}{2016}]%
        {covington2016deep}
\bibfield{author}{\bibinfo{person}{P. Covington}, \bibinfo{person}{J. Adams},
  {and} \bibinfo{person}{E. Sargin}.} \bibinfo{year}{2016}\natexlab{}.
\newblock \showarticletitle{Deep neural networks for {Youtube}
  recommendations}. In \bibinfo{booktitle}{\emph{RecSys}}.
\newblock


\bibitem[\protect\citeauthoryear{Craswell, Zoeter, Taylor, and Ramsey}{Craswell
  et~al\mbox{.}}{2008}]%
        {craswell2008experimental}
\bibfield{author}{\bibinfo{person}{Nick Craswell}, \bibinfo{person}{Onno
  Zoeter}, \bibinfo{person}{Michael Taylor}, {and} \bibinfo{person}{Bill
  Ramsey}.} \bibinfo{year}{2008}\natexlab{}.
\newblock \showarticletitle{An experimental comparison of click position-bias
  models}. In \bibinfo{booktitle}{\emph{WSDM 2008}}. \bibinfo{pages}{87--94}.
\newblock


\bibitem[\protect\citeauthoryear{D'Arcy}{D'Arcy}{2021}]%
        {marc_darcy_2021}
\bibfield{author}{\bibinfo{person}{Marc D'Arcy}.}
  \bibinfo{year}{2021}\natexlab{}.
\newblock \bibinfo{title}{Opinion: The 3 Post-COVID Trends Empowering People
  and Shaping the Future}.
\newblock
\newblock
\urldef\tempurl%
\url{https://adage.com/article/opinion/opinion-3-post-covid-trends-empowering-
  people-and-shaping-future/2342861}
\showURL{%
\tempurl}


\bibitem[\protect\citeauthoryear{Daulton et~al\mbox{.}}{Daulton
  et~al\mbox{.}}{2019}]%
        {daulton2019thompson}
\bibfield{author}{\bibinfo{person}{Samuel Daulton} {et~al\mbox{.}}}
  \bibinfo{year}{2019}\natexlab{}.
\newblock \showarticletitle{Thompson sampling for contextual bandit problems
  with auxiliary safety constraints}.
\newblock \bibinfo{journal}{\emph{arXiv:1911.00638}} (\bibinfo{year}{2019}).
\newblock


\bibitem[\protect\citeauthoryear{Daulton, Balandat, and Bakshy}{Daulton
  et~al\mbox{.}}{2021}]%
        {daulton2021nehvi}
\bibfield{author}{\bibinfo{person}{Samuel Daulton}, \bibinfo{person}{Maximilian
  Balandat}, {and} \bibinfo{person}{Eytan Bakshy}.}
  \bibinfo{year}{2021}\natexlab{}.
\newblock \showarticletitle{Parallel {Bayesian} Optimization of Multiple Noisy
  Objectives with Expected Hypervolume Improvement}, In
  \bibinfo{booktitle}{NeurIPS 34}.
\newblock \bibinfo{journal}{\emph{arXiv:2006.05078}}.
\newblock


\bibitem[\protect\citeauthoryear{Dickson}{Dickson}{2021}]%
        {dickson_2021}
\bibfield{author}{\bibinfo{person}{Ben Dickson}.}
  \bibinfo{year}{2021}\natexlab{}.
\newblock \bibinfo{title}{Why machine learning strategies fail}.
\newblock
\newblock
\urldef\tempurl%
\url{https://venturebeat.com/2021/02/25/why-machine-learning-strategies-fail/}
\showURL{%
\tempurl}


\bibitem[\protect\citeauthoryear{Dunn}{Dunn}{2016}]%
        {fblearner2016}
\bibfield{author}{\bibinfo{person}{Jeffrey Dunn}.}
  \bibinfo{year}{2016}\natexlab{}.
\newblock \bibinfo{title}{Introducing FBLearner Flow: Facebook’s AI
  backbone}.
\newblock
\newblock
\urldef\tempurl%
\url{https://engineering.fb.com/2016/05/09/core-data/introducing-fblearner-flow-facebook-s-ai-backbone}
\showURL{%
\tempurl}


\bibitem[\protect\citeauthoryear{Feng, Letham, Mao, and Bakshy}{Feng
  et~al\mbox{.}}{2020}]%
        {feng2020high}
\bibfield{author}{\bibinfo{person}{Qing Feng}, \bibinfo{person}{Benjamin
  Letham}, \bibinfo{person}{Hongzi Mao}, {and} \bibinfo{person}{Eytan Bakshy}.}
  \bibinfo{year}{2020}\natexlab{}.
\newblock \showarticletitle{High-dimensional contextual policy search with
  unknown context rewards using Bayesian optimization}.
\newblock \bibinfo{journal}{\emph{NeurIPS}}  \bibinfo{volume}{33}
  (\bibinfo{year}{2020}).
\newblock


\bibitem[\protect\citeauthoryear{Gauci et~al\mbox{.}}{Gauci
  et~al\mbox{.}}{2018}]%
        {gauci2018horizon}
\bibfield{author}{\bibinfo{person}{Jason Gauci} {et~al\mbox{.}}}
  \bibinfo{year}{2018}\natexlab{}.
\newblock \showarticletitle{Horizon: Facebook's Open Source Applied
  Reinforcement Learning Platform}.
\newblock \bibinfo{journal}{\emph{arXiv:1811.00260}} (\bibinfo{year}{2018}).
\newblock


\bibitem[\protect\citeauthoryear{Gupta et~al\mbox{.}}{Gupta
  et~al\mbox{.}}{2020}]%
        {Gupta2020FBDNN}
\bibfield{author}{\bibinfo{person}{Udit Gupta} {et~al\mbox{.}}}
  \bibinfo{year}{2020}\natexlab{}.
\newblock \showarticletitle{The Architectural Implications of Facebook's
  DNN-based Personalized Recommendation}.
\newblock \bibinfo{journal}{\emph{HPCA}} (\bibinfo{year}{2020}),
  \bibinfo{pages}{488--501}.
\newblock
\showeprint{1906.03109}


\bibitem[\protect\citeauthoryear{Hazelwood et~al\mbox{.}}{Hazelwood
  et~al\mbox{.}}{2018}]%
        {hazelwood2018applied}
\bibfield{author}{\bibinfo{person}{Kim Hazelwood} {et~al\mbox{.}}}
  \bibinfo{year}{2018}\natexlab{}.
\newblock \showarticletitle{Applied machine learning at {Facebook}: A
  datacenter infrastructure perspective}. In \bibinfo{booktitle}{\emph{HPCA
  2018}}. IEEE, \bibinfo{pages}{620--629}.
\newblock


\bibitem[\protect\citeauthoryear{Hermann and Del~Balso}{Hermann and
  Del~Balso}{2017}]%
        {Hermann2017Michelangelo}
\bibfield{author}{\bibinfo{person}{Jeremy Hermann} {and} \bibinfo{person}{Mike
  Del~Balso}.} \bibinfo{year}{2017}\natexlab{}.
\newblock \bibinfo{title}{Meet {M}ichelangelo: {U}ber’s {M}achine {L}earning
  {P}latform}.
\newblock
\newblock
\urldef\tempurl%
\url{https://eng.uber.com/michelangelo-machine-learning-platform/}
\showURL{%
\tempurl}


\bibitem[\protect\citeauthoryear{Kohavi, Longbotham, Sommerfield, and
  Henne}{Kohavi et~al\mbox{.}}{2009}]%
        {kohavi2009controlled}
\bibfield{author}{\bibinfo{person}{Ron Kohavi}, \bibinfo{person}{Roger
  Longbotham}, \bibinfo{person}{Dan Sommerfield}, {and}
  \bibinfo{person}{Randal~M Henne}.} \bibinfo{year}{2009}\natexlab{}.
\newblock \showarticletitle{Controlled experiments on the {Web}: survey and
  practical guide}.
\newblock \bibinfo{journal}{\emph{Data mining and knowledge discovery}}
  \bibinfo{volume}{18}, \bibinfo{number}{1} (\bibinfo{year}{2009}),
  \bibinfo{pages}{140--181}.
\newblock


\bibitem[\protect\citeauthoryear{Kraska et~al\mbox{.}}{Kraska
  et~al\mbox{.}}{2017}]%
        {kraska2017learned}
\bibfield{author}{\bibinfo{person}{Tim Kraska} {et~al\mbox{.}}}
  \bibinfo{year}{2017}\natexlab{}.
\newblock \showarticletitle{The Case for Learned Index Structures}.
\newblock \bibinfo{journal}{\emph{CoRR}} (\bibinfo{year}{2017}).
\newblock
\showeprint[arXiv]{1712.01208}


\bibitem[\protect\citeauthoryear{Künzel et~al\mbox{.}}{Künzel
  et~al\mbox{.}}{2019}]%
        {kunzel2019meta}
\bibfield{author}{\bibinfo{person}{Sören~R. Künzel} {et~al\mbox{.}}}
  \bibinfo{year}{2019}\natexlab{}.
\newblock \showarticletitle{Metalearners for estimating heterogeneous treatment
  effects using machine learning}.
\newblock \bibinfo{journal}{\emph{PNAS}} \bibinfo{volume}{116},
  \bibinfo{number}{10} (\bibinfo{date}{Feb} \bibinfo{year}{2019}),
  \bibinfo{pages}{4156–4165}.
\newblock
\showISSN{1091-6490}


\bibitem[\protect\citeauthoryear{Laud}{Laud}{2004}]%
        {laud2004reward}
\bibfield{author}{\bibinfo{person}{Adam~Daniel Laud}.}
  \bibinfo{year}{2004}\natexlab{}.
\newblock \bibinfo{booktitle}{\emph{Theory and application of reward shaping in
  reinforcement learning}}.
\newblock \bibinfo{publisher}{UIUC}.
\newblock


\bibitem[\protect\citeauthoryear{Li, Chu, Langford, and Schapire}{Li
  et~al\mbox{.}}{2010}]%
        {li2010contextual}
\bibfield{author}{\bibinfo{person}{Lihong Li}, \bibinfo{person}{Wei Chu},
  \bibinfo{person}{John Langford}, {and} \bibinfo{person}{Robert~E Schapire}.}
  \bibinfo{year}{2010}\natexlab{}.
\newblock \showarticletitle{A contextual-bandit approach to personalized news
  article recommendation}. In \bibinfo{booktitle}{\emph{WWW}}.
  \bibinfo{pages}{661--670}.
\newblock


\bibitem[\protect\citeauthoryear{Li et~al\mbox{.}}{Li et~al\mbox{.}}{2020}]%
        {pytorch2020}
\bibfield{author}{\bibinfo{person}{Shen Li} {et~al\mbox{.}}}
  \bibinfo{year}{2020}\natexlab{}.
\newblock \showarticletitle{PyTorch Distributed: Experiences on Accelerating
  Data Parallel Training}. In \bibinfo{booktitle}{\emph{VLDB}},
  Vol.~\bibinfo{volume}{13(12)}.
\newblock


\bibitem[\protect\citeauthoryear{Mao et~al\mbox{.}}{Mao et~al\mbox{.}}{2020}]%
        {mao2020real}
\bibfield{author}{\bibinfo{person}{Hongzi Mao} {et~al\mbox{.}}}
  \bibinfo{year}{2020}\natexlab{}.
\newblock \showarticletitle{Real-world video adaptation with reinforcement
  learning}.
\newblock \bibinfo{journal}{\emph{arXiv:2008.12858}} (\bibinfo{year}{2020}).
\newblock


\bibitem[\protect\citeauthoryear{Miranda}{Miranda}{2021}]%
        {miranda2021datacentric}
\bibfield{author}{\bibinfo{person}{Lester~James Miranda}.}
  \bibinfo{year}{2021}\natexlab{}.
\newblock \showarticletitle{Towards data-centric machine learning: a short
  review}.
\newblock  (\bibinfo{year}{2021}).
\newblock
\urldef\tempurl%
\url{https://ljvmiranda921.github.io/notebook/2021/07/30/data-centric-ml/}
\showURL{%
\tempurl}


\bibitem[\protect\citeauthoryear{Molino, Dudin, and Miryala}{Molino
  et~al\mbox{.}}{2019}]%
        {molino19ludwig}
\bibfield{author}{\bibinfo{person}{P. Molino}, \bibinfo{person}{Y. Dudin},
  {and} \bibinfo{person}{S.~S. Miryala}.} \bibinfo{year}{2019}\natexlab{}.
\newblock \showarticletitle{Ludwig: a type-based declarative deep learning
  toolbox}.
\newblock \bibinfo{journal}{\emph{arxiv:1909.07930}} (\bibinfo{year}{2019}).
\newblock


\bibitem[\protect\citeauthoryear{Molino and R\'{e}}{Molino and R\'{e}}{2021}]%
        {molino21declarative}
\bibfield{author}{\bibinfo{person}{P. Molino} {and} \bibinfo{person}{C.
  R\'{e}}.} \bibinfo{year}{2021}\natexlab{}.
\newblock \showarticletitle{Declarative Machine Learning Systems}.
\newblock \bibinfo{journal}{\emph{ACM Queue}}  \bibinfo{volume}{19}
  (\bibinfo{year}{2021}).
\newblock
Issue 3.


\bibitem[\protect\citeauthoryear{Naumov et~al\mbox{.}}{Naumov
  et~al\mbox{.}}{2019}]%
        {naumov2019dlrm}
\bibfield{author}{\bibinfo{person}{Maxim Naumov} {et~al\mbox{.}}}
  \bibinfo{year}{2019}\natexlab{}.
\newblock \showarticletitle{Deep Learning Recommendation Model for
  Personalization and Recommendation Systems}.
\newblock \bibinfo{journal}{\emph{CoRR}}  \bibinfo{volume}{abs/1906.00091}
  (\bibinfo{year}{2019}).
\newblock


\bibitem[\protect\citeauthoryear{Orr et~al\mbox{.}}{Orr et~al\mbox{.}}{2021}]%
        {featurestore2021}
\bibfield{author}{\bibinfo{person}{Laurel~J. Orr} {et~al\mbox{.}}}
  \bibinfo{year}{2021}\natexlab{}.
\newblock \showarticletitle{Managing {ML} Pipelines: Feature Stores and the
  Coming Wave of Embedding Ecosystems}.
\newblock \bibinfo{journal}{\emph{CoRR}} (\bibinfo{year}{2021}).
\newblock
\showeprint[arXiv]{2108.05053}


\bibitem[\protect\citeauthoryear{Paleyes, Urma, and Lawrence}{Paleyes
  et~al\mbox{.}}{2020}]%
        {paleyes2020challenges}
\bibfield{author}{\bibinfo{person}{Andrei Paleyes},
  \bibinfo{person}{Raoul-Gabriel Urma}, {and} \bibinfo{person}{Neil~D
  Lawrence}.} \bibinfo{year}{2020}\natexlab{}.
\newblock \showarticletitle{Challenges in deploying machine learning: a survey
  of case studies}.
\newblock \bibinfo{journal}{\emph{arXiv:2011.09926}} (\bibinfo{year}{2020}).
\newblock


\bibitem[\protect\citeauthoryear{R{\'e} et~al\mbox{.}}{R{\'e}
  et~al\mbox{.}}{2019}]%
        {re2019overton}
\bibfield{author}{\bibinfo{person}{Christopher R{\'e}} {et~al\mbox{.}}}
  \bibinfo{year}{2019}\natexlab{}.
\newblock \showarticletitle{Overton: A data system for monitoring and improving
  machine-learned products}.
\newblock \bibinfo{journal}{\emph{arXiv:1909.05372}} (\bibinfo{year}{2019}).
\newblock


\bibitem[\protect\citeauthoryear{Rodr{\'\i}guez, Bautista, Gonz{\`a}lez, and
  Escalera}{Rodr{\'\i}guez et~al\mbox{.}}{2018}]%
        {rodriguez2018beyond}
\bibfield{author}{\bibinfo{person}{Pau Rodr{\'\i}guez},
  \bibinfo{person}{Miguel~A Bautista}, \bibinfo{person}{Jordi Gonz{\`a}lez},
  {and} \bibinfo{person}{Sergio Escalera}.} \bibinfo{year}{2018}\natexlab{}.
\newblock \showarticletitle{Beyond One-hot Encoding: lower dimensional target
  embedding}.
\newblock \bibinfo{journal}{\emph{arXiv:1806.10805}} (\bibinfo{year}{2018}).
\newblock


\bibitem[\protect\citeauthoryear{Roy}{Roy}{2016}]%
        {fblite2016}
\bibfield{author}{\bibinfo{person}{Gautam Roy}.}
  \bibinfo{year}{2016}\natexlab{}.
\newblock \bibinfo{title}{How we built Facebook Lite for every Android phone
  and network}.
\newblock
\newblock
\urldef\tempurl%
\url{https://engineering.fb.com/2016/03/09/android/how-we-built-facebook-lite-for-every-android-phone-and-network}
\showURL{%
\tempurl}


\bibitem[\protect\citeauthoryear{Sagar}{Sagar}{2021}]%
        {sagar_2021}
\bibfield{author}{\bibinfo{person}{Ram Sagar}.}
  \bibinfo{year}{2021}\natexlab{}.
\newblock \bibinfo{title}{Andrew Ng Urges ML Community To Be More
  Data-Centric}.
\newblock
\newblock
\urldef\tempurl%
\url{https://analyticsindiamag.com/big-data-to-good-data-andrew-ng-urges-ml-community-to-be-more-data-centric-and-less-model-centric/}
\showURL{%
\tempurl}


\bibitem[\protect\citeauthoryear{Sambasivan et~al\mbox{.}}{Sambasivan
  et~al\mbox{.}}{2021}]%
        {sambasivan2021everyone}
\bibfield{author}{\bibinfo{person}{Nithya Sambasivan} {et~al\mbox{.}}}
  \bibinfo{year}{2021}\natexlab{}.
\newblock \showarticletitle{"Everyone wants to do the model work, not the data
  work": Data Cascades in High-Stakes AI}.
\newblock \bibinfo{journal}{\emph{SIGCHI, ACM}} (\bibinfo{year}{2021}).
\newblock


\bibitem[\protect\citeauthoryear{Sculley et~al\mbox{.}}{Sculley
  et~al\mbox{.}}{2015}]%
        {sculley2015hidden}
\bibfield{author}{\bibinfo{person}{David Sculley} {et~al\mbox{.}}}
  \bibinfo{year}{2015}\natexlab{}.
\newblock \showarticletitle{Hidden technical debt in machine learning systems}.
\newblock \bibinfo{journal}{\emph{NIPS}}  \bibinfo{volume}{28}
  (\bibinfo{year}{2015}), \bibinfo{pages}{2503--2511}.
\newblock


\bibitem[\protect\citeauthoryear{Soifer et~al\mbox{.}}{Soifer
  et~al\mbox{.}}{2019}]%
        {soifer2019inference}
\bibfield{author}{\bibinfo{person}{Jonathan Soifer} {et~al\mbox{.}}}
  \bibinfo{year}{2019}\natexlab{}.
\newblock \showarticletitle{Deep Learning Inference Service at Microsoft}. In
  \bibinfo{booktitle}{\emph{{USENIX} Conf. Operational Machine Learning
  (OpML)}}. \bibinfo{publisher}{{USENIX}}, \bibinfo{pages}{15--17}.
\newblock
\urldef\tempurl%
\url{https://www.usenix.org/conference/opml19/presentation/soifer}
\showURL{%
\tempurl}


\bibitem[\protect\citeauthoryear{Stein}{Stein}{2019}]%
        {stein2019}
\bibfield{author}{\bibinfo{person}{Gregory~J. Stein}.}
  \bibinfo{year}{2019}\natexlab{}.
\newblock \bibinfo{title}{Proxy metrics are everywhere in machine learning}.
\newblock
\newblock
\urldef\tempurl%
\url{http://cachestocaches.com/2019/1/proxy-metrics-are-everywhere-machine-lea}
\showURL{%
\tempurl}


\bibitem[\protect\citeauthoryear{Vartak and Madden}{Vartak and Madden}{2018}]%
        {vartak2018modeldb}
\bibfield{author}{\bibinfo{person}{M. Vartak} {and} \bibinfo{person}{S.
  Madden}.} \bibinfo{year}{2018}\natexlab{}.
\newblock \showarticletitle{MODELDB: Opportunities and Challenges in Managing
  Machine Learning Models.}
\newblock \bibinfo{journal}{\emph{IEEE Data Eng. Bull.}} \bibinfo{volume}{41},
  \bibinfo{number}{4} (\bibinfo{year}{2018}), \bibinfo{pages}{16--25}.
\newblock


\bibitem[\protect\citeauthoryear{Wager and Athey}{Wager and Athey}{2018}]%
        {wager2018estimation}
\bibfield{author}{\bibinfo{person}{S. Wager} {and} \bibinfo{person}{S. Athey}.}
  \bibinfo{year}{2018}\natexlab{}.
\newblock \showarticletitle{Estimation and inference of heterogeneous treatment
  effects using random forests}.
\newblock \bibinfo{journal}{\emph{J. Amer. Stat. Assoc.}}
  \bibinfo{volume}{113}, \bibinfo{number}{523} (\bibinfo{year}{2018}),
  \bibinfo{pages}{1228--1242}.
\newblock


\bibitem[\protect\citeauthoryear{Wang, Wang, and Ma}{Wang
  et~al\mbox{.}}{2019}]%
        {wang2019predictive}
\bibfield{author}{\bibinfo{person}{Hanson Wang}, \bibinfo{person}{Zehui Wang},
  {and} \bibinfo{person}{Yuanyuan Ma}.} \bibinfo{year}{2019}\natexlab{}.
\newblock \showarticletitle{Predictive Precompute with Recurrent Neural
  Networks}.
\newblock \bibinfo{journal}{\emph{arXiv:1912.06779}} (\bibinfo{year}{2019}).
\newblock


\bibitem[\protect\citeauthoryear{Wu et~al\mbox{.}}{Wu et~al\mbox{.}}{2021}]%
        {wu2021sustainable}
\bibfield{author}{\bibinfo{person}{Carole-Jean Wu} {et~al\mbox{.}}}
  \bibinfo{year}{2021}\natexlab{}.
\newblock \bibinfo{title}{Sustainable AI: Environmental Implications,
  Challenges and Opportunities}.
\newblock
\newblock
\showeprint[arxiv]{2111.00364}~[cs.LG]


\bibitem[\protect\citeauthoryear{Xu et~al\mbox{.}}{Xu et~al\mbox{.}}{2015}]%
        {xu2015infrastructure}
\bibfield{author}{\bibinfo{person}{Ya Xu} {et~al\mbox{.}}}
  \bibinfo{year}{2015}\natexlab{}.
\newblock \showarticletitle{From infrastructure to culture: A/B testing
  challenges in large scale social networks}. In
  \bibinfo{booktitle}{\emph{KDD}}. \bibinfo{pages}{2227--2236}.
\newblock


\bibitem[\protect\citeauthoryear{Yankov, Berkhin, and Li}{Yankov
  et~al\mbox{.}}{2015}]%
        {yankov2015evaluation}
\bibfield{author}{\bibinfo{person}{Dragomir Yankov}, \bibinfo{person}{Pavel
  Berkhin}, {and} \bibinfo{person}{Lihong Li}.}
  \bibinfo{year}{2015}\natexlab{}.
\newblock \showarticletitle{Evaluation of explore-exploit policies in
  multi-result ranking systems}.
\newblock \bibinfo{journal}{\emph{arXiv:1504.07662}} (\bibinfo{year}{2015}).
\newblock


\bibitem[\protect\citeauthoryear{Zhao et~al\mbox{.}}{Zhao
  et~al\mbox{.}}{2019}]%
        {zhao2019recommending}
\bibfield{author}{\bibinfo{person}{Zhe Zhao} {et~al\mbox{.}}}
  \bibinfo{year}{2019}\natexlab{}.
\newblock \showarticletitle{Recommending what video to watch next: a multitask
  ranking system}. In \bibinfo{booktitle}{\emph{RecSys `19}}.
  \bibinfo{pages}{43--51}.
\newblock


\end{thebibliography}

\newpage

\begin{figure}[t]
%\vspace{-4mm}
  \centering
  \includegraphics[width=0.82\linewidth]{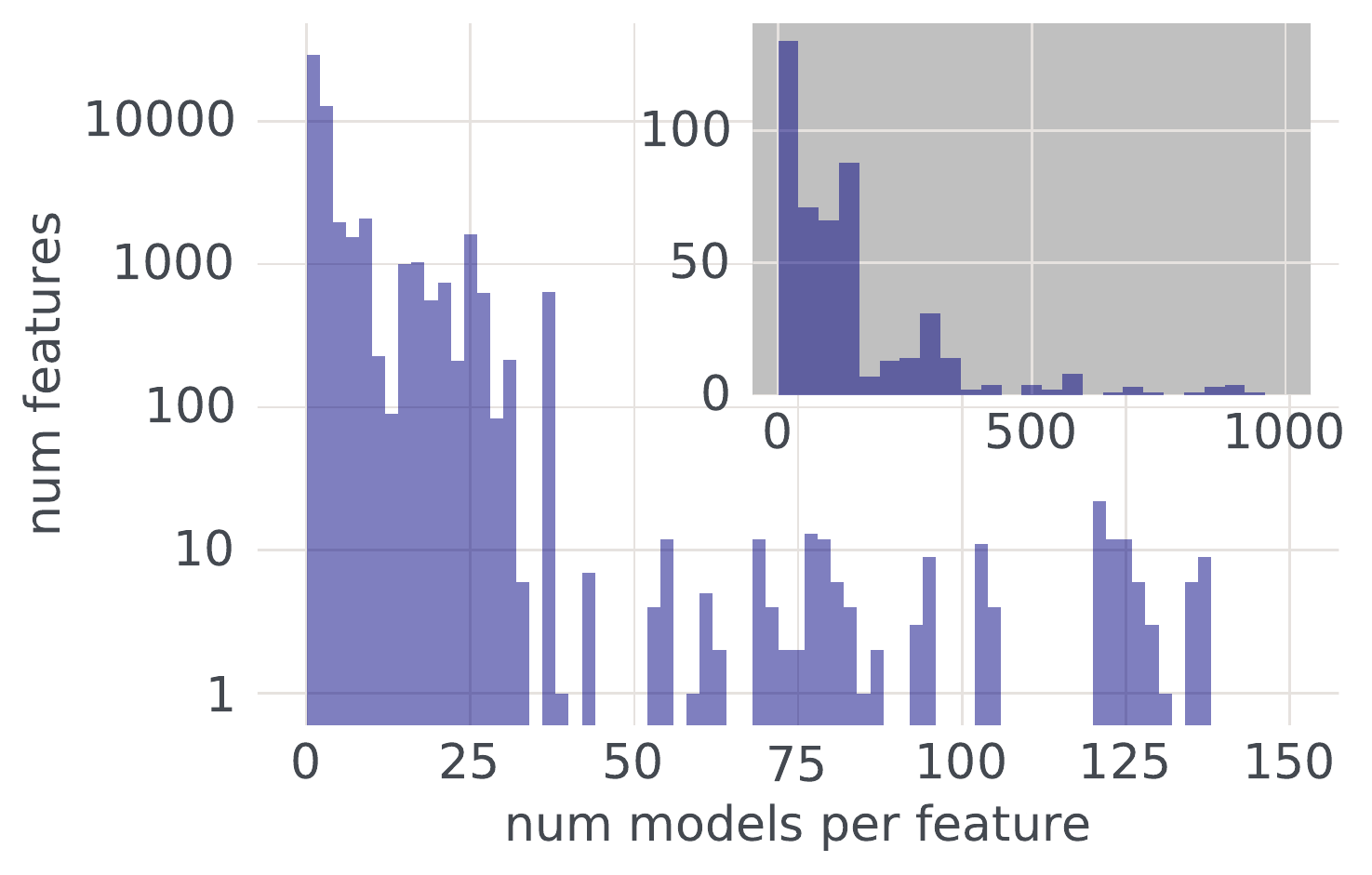}
 %\vspace{-1mm} 
\caption{
\label{fig:features}
 Histograms showing how many \Looper models use a given feature (outer) and features per model (inner).}
\vspace{-2mm}
% https://www.internalfb.com/intern/anp/view/?id=562915
\end{figure}

\appendix
\section{Resource utilization and its optimization}
\label{sec:resources}

\begin{figure}[t]
%\vspace{-4mm}
  \centering
  \includegraphics[width=0.9\linewidth]{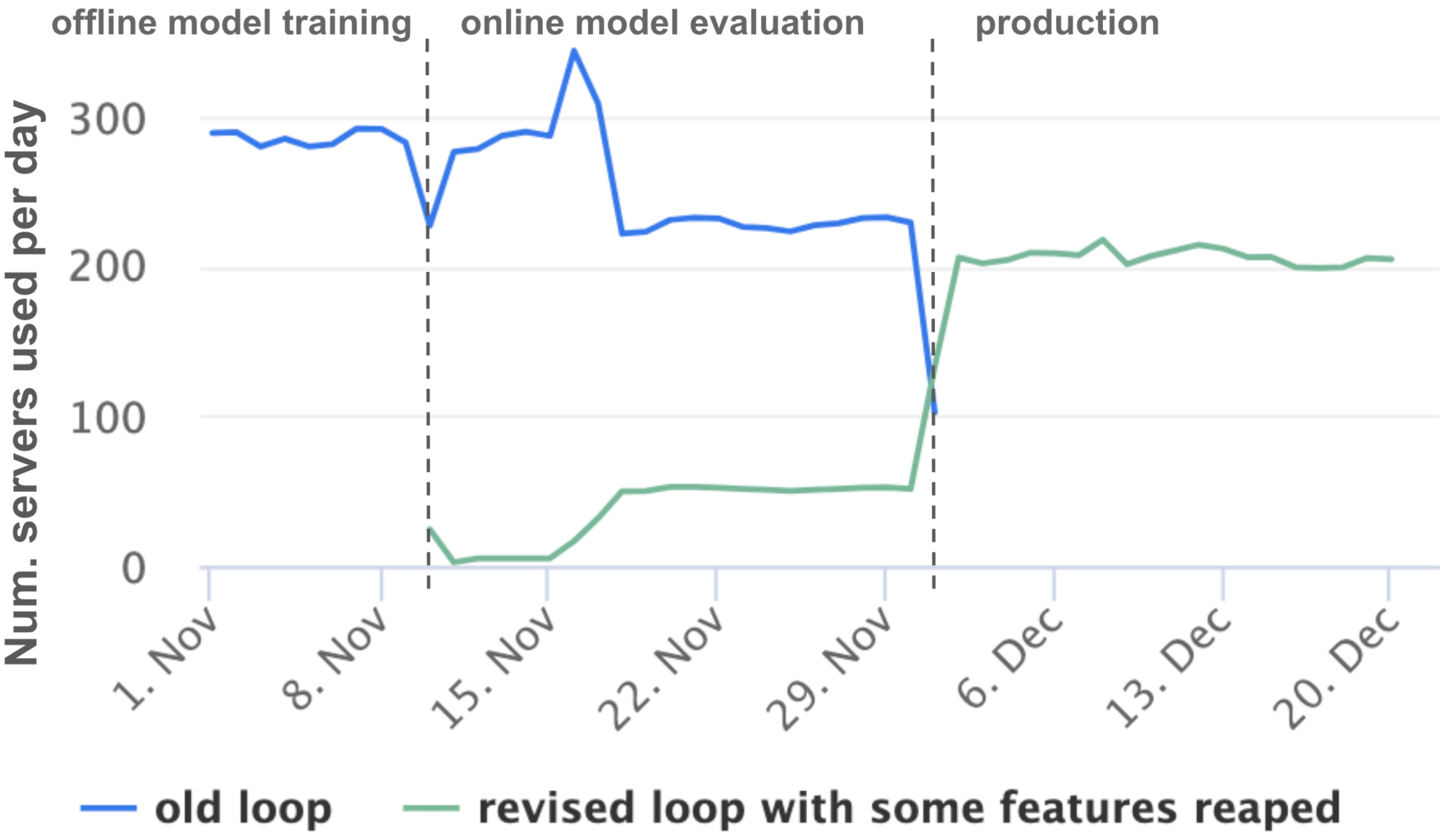}
 \vspace{-1mm} 
\caption{
\label{fig:resource_optimization}
 The three phases of resource optimization via {\em feature reaping}. \\ Resource consumption is decreased without adverse product impacts.}
\vspace{-4mm}
% https://www.internalfb.com/intern/anp/view/?id=562915
\end{figure}

 Smart strategies tend to provide significant benefits but sometimes need serious computational resources,\eat{\footnote{See resource utilization of various ML models in {\tt https://openai.com/blog/ai-and-compute/} } } Therefore
 product deployment requires judicious resource management.
\hush{Cumulatively, \Looper currently uses 2-3 thousand production servers.}
%~\cite{hazelwood2018applied}. 
\Looper is deployed in numerous and diverse applications at \Meta, some of which optimize performance of other systems and some enhance functionality (Figure \ref{fig:use_case_type}). \hush{This makes it difficult to report overall trends for optimizing resource utilization, but} Such economies-of-scale infrastructure enables resource reuse and load-balancing.
Figure \ref{fig:use_case_resource_type} shows that different use cases exhibit different model-lifecycle bottlenecks, with {\em feature extraction} drawing the largest share of resources for demanding use cases.\footnote{We note that \Looper deploys moderate-complexity models with {\em diverse metadata features}, whereas advanced deep learning models with {\em homogeneous} image pixels, word embeddings, etc may exhibit different trends.} Based on this trend, we developed a {\em feature reaping} optimization that removes unimportant features. This optimization estimates the importance of individual features, removes features in groups, then checks the results by training a reduced model and evaluating it. When we deployed feature reaping in production, it and provided overall 11\% resource-cost savings (10-30\% per use case) with no adverse product impacts. Figure \ref{fig:resource_optimization} illustrates the resource savings (28\%) for one use case via feature reaping. It distinguishes 3 stages that experience constant decision traffic: ($i$) offline model training that runs feature reaping and trains a reduced model, ($ii$) the online stage that uses the old and the revised (20\% traffic only) model to evaluate product impact, and ($iii$) the production stage that uses only the new model whose performance had been validated by online evaluation (which took 2 weeks to collect statistically significant results). Sharing such AutoML  optimizations
across multiple use cases makes our platform competitive with specialized platforms.
%{As \Looper scales to more use cases and greater %model complexity, trading model performance for %resource metrics is growing in importance.

\begin{figure}[hb]
  \centering
  \includegraphics[width=0.90\linewidth]{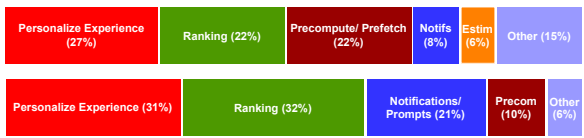}
%  \vspace{-1mm}                                     
  \caption{\label{fig:use_case_type} Product adoption of smart strategies by use case count (top) and by resource consumption (bottom).
  Resource consumption during training is correlated with data amount, but resource consumption at inference reflects decision rates in applications,
  the number of features used by models and the presence of synthetic/engineered features.
  }
  \Description{}
% \vspace{-4mm}
\end{figure}

\section{A survey of platform adopters}
\label{sec:survey}
%  https://fb.quip.com/bYz0AXl7kF0i

%Embedding such smart strategies into modern systems \hush{must be driven by application/product  metrics, how the system is structured, what type of decisions can be improved, and what data is available for these improvements. It is also critical to find}
%requires appropriate entry points in the software structure and the software engineering process --- the lack of a good entry point can be a showstopper, whereas an apt entry point can minimize the adoption burden. 
The article “Why Machine Learning Strategies Fail”~\cite{dickson_2021} lists common barriers to entry:
\begin{itemize}
\item lacking a business case,
\item lacking data,
\item lacking ML talent,
\item lacking sufficient in-house ML expertise for outsourcing,
\item failing to evaluate an ML strategy.
\end{itemize}
When talking to prospective clients, we advise against using \Looper when there is no need for an end-to-end platform. For example, in a Kaggle-like environment with well-defined data,
the goal is to train a model to optimize closed-form objectives for model performance. A second example is tasks without a clear product metric, such as building latent-space embeddings and other self-supervised tasks. Yet another reason not to use \Looper is the efficiency of special-case platforms, e.g., for ranking and image-processing.

To clarify when \Looper is relevant and to clarify the adoption process of smart strategies, we interviewed product teams at \Meta that adopted our platform and saw product impacts (Figure \ref{fig:use_case_type}).\hush{These teams pursued optimized notification delivery, personalized UI experience, ranking products, and prefetching media to mobile clients. We asked about the team’s motivation to adopt ML, their ML expertise, the thought process on choosing among several available ML systems, the actual impact, the overall timeline and the amount of effort required, as well as the pitfalls and barriers to entry they ran into.} All the teams had tried heuristic approaches but with poor results, hence their focus on ML. Simple heuristics proved insufficient for user bases spanning multiple countries with distinct demographic and usage patterns. The following challenges were highlighted:
%\begin{itemize}
\circled{1}
manually optimizing parameters in large search spaces,
\circled{2}
figuring out the correct rules to make heuristics effective,
\circled{3}
trading off multiple objectives,
\circled{4}
updating heuristic logic quickly, especially in on-device code.

The spectrum of ML expertise varied across product teams from beginners to experienced ML engineers, and only 15\% of teams using our platform include ML engineers.
\hush{Regardless of experience, all adopters noted the benefits of a comprehensive {\em product-driven} smart-strategies platform.}
For teams without production ML experience, an easy-to-use ML platform is often the deciding factor for ML adoption, and ML investment continues upon evidence of utility. An engineer mentioned that a lower-level ML system had a confusing development flow and unwieldy debugging. They were unable to set up recurring model training and publishing. \Looper abstracts away concerns about SW upgrades, logging, monitoring, etc behind high-level services and unlocks hefty productivity savings. 

For experienced ML engineers, a smart-strategies platform improves productivity by automating repetitive time-consuming work: writing database queries, implementing data pipelines, setting up monitoring and alerts. Compared to\hush{less-integrated} narrow-focus systems, it helps product developers launch more ML use cases. An engineer shared prior experience 
writing custom queries for features and labels, and manually setting up pipelines for recurring training and model publishing without an easy way to monitor model performance and issue emergency alerts.
Some prospective clients who evaluated our platform chose other ML platforms
within our company or stayed with their custom-designed infrastructure. They 
%mentioned %the following omissions in our platform: 
missed batched offline prediction with mega-sized data\hush{() automatic tuning of sampling rates for
training data with extreme class imbalance,}
%($ii$) custom provisions for long retention of training data, 
%($ii$)
and needed exceptional performance
possible only with custom ML models.
%noted that generic ML models on our
%platform (slightly) underperformed custom ML models.
These issues can be addressed with additional platform development efforts.
%Some product teams evaluated our support for personalization but found no benefits.

%Our product-oriented smart-strategies platform is built for ease of use. 
 Successful platform adopters configured initial ML models in two days and started collecting training data. Training the model using product feedback and revising it over 1-2 weeks enabled online product experiments that take 2-4 weeks.\hush{, depending on data volume, metric sensitivity, and organizational norms.} 
 \hush{An ML model can be launched to production when empirical results meet launch criteria, otherwise the team revises the model and runs more experiments. }
 HTE analysis and impact optimization take 1-3 weeks. Product launch can occur 1-3 months after initial data collection. 
Among platform adopters, experienced engineers aware of ML-related technical debt and risks~\cite{sculley2015hidden,agarwal2016making,paleyes2020challenges,dickson_2021,sambasivan2021everyone} appreciated the built-in support for recurring training, model publishing, data visualization, as well as
monitoring label and feature distributions over time with data-drift alerts.
\hush{in production.} Also noted was the canarying mechanism for new models (Section \ref{sec:core}).
Surprisingly important was helping adopters share model insights (such as feature importance) with leadership and product managers. Among possible improvements, adopters mentioned development velocity.

\section{Improving product-development velocity}
\hush{
\cite{rai2020logic}
\cite{lundberg2020}
\cite{ribeiro2016should}
\cite{sculley2015hidden}
\cite{agrawal2019data}
}
As a platform used primarily by product engineers, Looper aims to reduce friction while users modify parameters and model configurations. This is accomplished in two ways: 
\begin{enumerate}
\item a GUI to easily edit all Looper parameters, and
\item automatic optimization of configuration parameters.
\end{enumerate}

Every aspect of configuration is represented in the UI. Each use case has a landing page to manage all related models, data versions, and experiments. All parameters including features, labels, and current production models are displayed in an editable form for users to modify and immediately push changes to production. Experiments can be launched from the use case landing page and the UI allows users to select baseline and experiment models to immediately compare them using integrated A/B tests.

All parameters are stored in configuration and this enables automatic optimization of parameters. Using the integrated experiment optimization system \cite{bakshy2018ae}, several values for a parameter are optimally selected to immediately test in our integrated A/B experimentation platform. These values can then be tested against product metrics. Since most parameters required for experimentation are already stored in strategy blueprints, including eventual support for product metrics, it is possible to automatically improve various parameter values using the experiment optimization system and A/B test to check if the config optimization benefits product, system, and resource usage metrics without affecting production models and without use case owner input.

\section{Development effort}
The \Looper platform described in this paper
was developed at \Meta over several years. It uses
several types of software infrastructure, such as databases, horizontal ML platforms, reusable ML models and frameworks, product metrics, and support for product A/B testing. The overall design was revised to better adapt to the needs of applications. 
On the other hand, a team of ten experienced software engineers should be able to implement our core platform design in half a year using relevant open-source and/or company infrastructure. In such development, it is important to focus on representative product use cases and guide software development within a well-defined scope. Avoiding common problems, rather than developing comprehensive solutions, can reduce time to first application. In particular, the complete chain of custody of data in \Looper helps avoid or reduce many common problems with data quality, such as delayed and missing data, mismatches between training and testing, etc. 

\end{document}